\documentclass[12pt]{article}

\usepackage{sbc-template}
\usepackage{graphicx,url}
\usepackage[utf8]{inputenc}
\usepackage{url}
\usepackage{balance}
\usepackage[normalem]{ulem}
\useunder{\uline}{\ul}{}
\usepackage{supertabular,booktabs}
\usepackage{tabularx}
\usepackage{longtable}
\usepackage{graphicx}
\usepackage{xcolor}
\usepackage{adjustbox}
\usepackage{multirow}
\usepackage{makecell}
\usepackage{longtable}
\usepackage{supertabular}
\usepackage{tabularx, booktabs, longtable}
\usepackage[labelformat=simple]{subcaption}

\usepackage{etoolbox}
\newcolumntype{C}{>{\centering\arraybackslash}X} % centered version of "X" type
\usepackage[acronym]{glossaries}
\newacronym{mtl}{MTL}{Multi-task Learning}
\newacronym{gan}{GAN}{Generative Adversarial Network}
\newacronym{fcn}{FCN}{Fully Convolutional Network}
\newacronym{gam}{GAM}{Gate Attention Module}
\newacronym{dam}{DAM}{Decoder Attention Module}
\newacronym{rab}{RAB}{Residual Attention Block}
\newacronym{ra}{RA}{Reverse Attention}
\newacronym{idc}{IDC}{Improved Dilation Convolution}
\newacronym{ag}{AG}{Attention Gate}
\newacronym{tasegnet}{TA-SegNet}{Tri-level Attention-based Segmentation Network}
\newacronym{tau}{TAU}{Tri-level Attention Unite}
\newacronym{qapnet}{QAP-Net}{Quadruple Augmented Pyramid Network}
\newacronym{raiunet}{RAIU-Net}{Residual Attention Inception U-Net}
\newacronym{chsnet}{CHS-Net}{COVID-19 Hierarchical Segmentation Network}
\newacronym{csse}{CSSE}{Systems Science and Engineering}
\newacronym{jhu}{JHU}{Johns Hopkins University}
\newacronym{rtpcr}{RT-PCR}{Reverse-Transcription Polymerase Chain Reaction}
\newacronym{ct}{CT}{Computed Tomography}
\newacronym{fpn}{FPN}{Feature Pyramid Network}
\newacronym{pspnet}{PSPNet}{Pyramid Scene Parsing Network}
\newacronym{manet}{MA-Net}{Multi-scale Attention Net}
\newacronym{se}{SE-Net}{Squeeze-and-Excitation Network}
\newacronym{ggo}{GGO}{Ground Glass Opacity}
\newacronym{iou}{IoU}{Intersection over Union}
\newacronym{rbc}{RBC}{Random Brightness Contrast}
\newacronym{clahe}{CLAHE}{Contrast Limited Adaptive Histogram Equalization}
\newacronym{sod}{SOD}{Salient Object Detection}
\newacronym{da}{DA}{Data Augmentation}
\title{Light In The Black: An Evaluation of Data Augmentation Techniques for COVID-19 CT's Semantic Segmentation}

\author{
        Bruno A. Krinski, Daniel V. Ruiz, and Eduardo Todt
        } 

\address{
  Department of Informatics\\
   Federal University of Paran\'a (UFPR) -- Curitiba, PR -- Brazil
  \email{\textit{\{bakrinski, dvruiz, todt\}@inf.ufpr.br}}
}

\begin{document} 

\maketitle

\begin{abstract}
    With the COVID-19 global pandemic, computer-assisted diagnoses of medical images have gained much attention, and robust methods of Semantic Segmentation of Computed Tomography~(CT) became highly desirable. Semantic Segmentation of CT is one of many research fields of automatic detection of COVID-19 and has been widely explored since the COVID-19 outbreak. In this work, we propose an extensive analysis of how different data augmentation techniques improve the training of encoder-decoder neural networks on this problem. Twenty different data augmentation techniques were evaluated on five different datasets. Each dataset was validated through a five-fold cross-validation strategy, thus resulting in over 3,000 experiments. Our findings show that spatial level transformations are the most promising to improve the learning of neural networks on this problem.
\end{abstract}
     
\begin{resumo} 
  Com a COVID-19, diagnósticos de imagens médicas assistidos por computador ganharam muita atenção, e métodos robustos de Segmentação Semântica de Tomografia Computadorizada (TC) tornaram-se altamente desejáveis. A Segmentação Semântica de TC é um dos muitos campos de pesquisa de detecção automática da COVID-19 e foi amplamente explorado desde o surto da COVID-19. Neste trabalho, propomos uma análise extensiva sobre o quanto diferentes técnicas de aumento de dados contribuem para melhorar o treinamento de redes neurais codificador-decodificador sobre este problema. Vinte técnicas diferentes de aumento de dados foram avaliadas em cinco conjuntos de dados diferentes. Cada conjunto de dados foi validado através de uma estratégia de validação cruzada de cinco subconjuntos, resultando assim em mais de 3.000 experimentos. Nossas descobertas mostram que as transformações de nível espacial são as mais promissoras para melhorar o aprendizado das redes neurais sobre este problema.
\end{resumo}

%%%%%%%%%%%%%%%%%%%%%%%%%%%%%%%%%%%%%%%%%%%%%%%%%%%%%%%%%%%%%%%%%%%%%%%%%%%%%%%%
\section{Introduction}

% \textbf{falar do covid e o numero de infectados.  E como a visão computacional pode auxiliar no diagnostico em tomografias.}

Since 2019 the world has struggled with the new coronavirus (COVID-19) pandemic, with millions of infections and deaths worldwide~\cite{Wang2020}. Until now, there are a total of 427,169,421 global cases and a total of 5,902,878 global deaths~\cite{coviddata} (updated February 22th, 2022). Due to the virus's quick dissemination, early diagnosis is highly desirable for faster treatment and tracking infected people~\cite{Chen2020.04.06.20054890}. Automatic detection of COVID-19 infections in \glspl*{ct} shows to be a great help for early diagnoses~\cite{Shi2021}, with the Semantic Segmentation~\cite{Cao2020} of \glspl*{ct} being widely explored since the COVID-19 outbreak~\cite{Shi2021}. Deep Learning based techniques and Deep Neural Networks achieved impressive results in the segmentation of COVID-19 \glspl*{ct}~\cite{Shi2021, Krinski2021}. However, it has two limiting factors. The first one is that labeling Semantic Segmentation is a labor-intensive and timing-consuming process, and each pixel of the image must receive the correct label. Otherwise, the network could converge to wrong results~\cite{Shi2021,Cao2020}. In addition to that, labeling \gls*{ct} segmentation datasets must be made by highly specialized doctors to properly label the lesion regions of the image~\cite{Shi2021}. 

With new approaches being proposed quickly, an urgency aggravated by the global pandemic, the need for a proper evaluation becomes apparent. A broad benchmark of architectures was presented by~\cite{Krinski2021}, and one of their conclusions was that the models' generalization was impaired by the small number of samples on the field's datasets which also suffer from class imbalance introducing some bias to the models. Data augmentation can mitigate this issue; however, the influence of data augmentation during training was left out. 

In this work, we propose an extensive analysis of how different data augmentation techniques improve the training of encoder-decoder neural networks on this problem. Twenty different data augmentation techniques were evaluated, see section~\ref{sec:exp}, in three distinct experiments using five \glspl*{ct} datasets:  MedSeg~\cite{medseg}, Zenodo~\cite{zenodo}, CC-CCII~\cite{Zhang2020}, MosMed~\cite{Morozov2020}, Ricord1a~\cite{ricord1a}. Each dataset was validated through a five-fold cross-validation strategy, thus resulting in over 3,000 experiments. The code for running these same experiments is publicly available\footnote{https://github.com/VRI-UFPR/SparkInTheDarkLars2021}.

% \textbf{falar dos datasets e experimentos brevemente para retormar na conclusao}

%%%%%%%%%%%%%%%%%%%%%%%%%%%%%%%%%%%%%%%%%%%%%%%%%%%%%%%%%%%%%%%%%%%%%%%%%%%%%%%%

\section{Related Work}
% Usually, these techniques add new information to the image and generate a new one.
Data augmentation aims to generate a synthetic image by applying different operations to a preexisting labeled image~\cite{ruiz2020giraffes}. The most common operation are variations of an affine transformation such as flip, translate, rotate, scale. The Random Erasing~\cite{Zhong2020} is one example of data augmentation that adds information to the original image. In this technique, a rectangle with random values is positioned in the image. This rectangle's height, width, and center points are random values. This data augmentation helps the network learning process be more robust to object occlusions. Also, it reduces the overfitting in the training step. The CutMix~\cite{Yun2019} follows the same idea of the Random Erasing. However, instead of using constant values or even random ones, the technique mixes two images by adding an image A to some region of image B. This reduces information loss and encourages generalization.

% However, it mixes two images by adding one image A to some region of image B. The advantage over those that fill a rectangle region of the image with zeros or random noise values is that it reduces the information loss, which grows the training efficiency.

Following the same line of CutMix~\cite{Yun2019}, the study presented by~\cite{Summers2019} evaluated several different non-linear mixing algorithms. The authors showed that non-linear mixing algorithms are also effective as linear mixing data augmentations. \cite{hendrycks2020augmix} proposed a mixing data augmentation called AugMix. In AugMix, sequences of data augmentations are applied in parallel, generating a different image for each data augmentation sequence. In the end, an element-wise convex combination is applied to mix all generated images. In~\cite{Kisantal2019}, the authors proposed a data augmentation technique for small objects to improve such objects’ detection and segmentation. The authors applied "copy and pasting" strategy to create several copies of the objects of interest. They showed that this strategy increases the number of anchor boxes generated by the Mask-RCNN~\cite{He2017}, which helps the network to learn and detect small objects. The ANDA~\cite{ruiz2019anda} and IDA \cite{ruiz2020ida} techniques follow the idea of introducing new objects, however since those are techniques focused on the generic problem of \gls{sod}, some additional operations are necessary such as Image Inpainting to erase the original object and some additional computation to choose which combination of background and object produce a significant salience and the affine transformations to be applied to the new object that will replace the original one.

In the Grid Mask~\cite{chen2020gridmask}, a mask composed of black squares uniformly distributed is generated and applied on top of the image. This augmentation follows the same idea of Random Erasing, which forces the network to learn from occluded regions in the image and reduces overfitting. The advantage over others that randomly remove regions from the image is that random algorithms can remove relevant regions from the image. The InstaBoost~\cite{fang2019instaboost} uses an inpainting technique~\cite{990497} to remove the interest object from the image and place it in another region in the image. An appearance consistency heatmap~\cite{Field1993} is used to estimate the new region of the image where the object will be placed. 

The authors of~\cite{Liu2019} proposed a data augmentation based on image-to-image divided into two steps: training and deployment. The proposed method uses several images of different classes in the training step, called source images, and learns to translate the images between these classes. Then, in the deployment step, a small set of images from the target class is used, with the proposed method being able to translate from the source classes to the target class. For each image to be mixed in the SuperMix~\cite{dabouei2020}, a mixing binary mask with the salience information of the respective image is generated. Then, a teacher model already trained in the problem is applied to mix the images, optimize the position of the salient region in the mixed image, and ensure that both salient regions are presented in the final mixed image. 

These \glspl*{da} are, in general, for generic segmentation problems. However, there is no proper comparison of \gls*{da} methods for the COVID-19 segmentation problem. In this work, we focus on performing a extensive comparison of generic \gls*{da} methods in the approached problem. 

% \textbf{escrever em um paragrafo o que difere dos trabalhos relacionados}

%%%%%%%%%%%%%%%%%%%%%%%%%%%%%%%%%%%%%%%%%%%%%%%%%%%%%%%%%%%%%%%%%%%%%%%%%%%%%%%%
\section{Metrics and Datasets}

% \textbf{juntar os dois primeiros paragrafos}

The models were trained and evaluated across five different \glspl*{ct} datasets: MedSeg~\cite{medseg}, Zenodo~\cite{zenodo}, CC-CCII~\cite{Zhang2020}, MosMed~\cite{Morozov2020}, and Ricord1a~\cite{ricord1a}. The MedSeg has 929 images and labels for four classes, with the following pixel proportion: Background (0.98563), \gls*{ggo} (0.01072), Consolidation (0.00351), and Pleural Effusion (0.0001). The Zenodo dataset has 3,520 images and labels for four classes, with the following pixel proportion: Background (0.89893), Left Lung (0.04331), Right Lung (0.04923), and Infections (0.00852). The MosMed dataset is composed of 2,049 images, with labels for two classes, with the following pixel proportion: Background (0.99810) and \gls*{ggo}-Consolidation (0.00189). Ricord dataset is divided into three sets: 1a, 1b, and 1c. The set 1a is the only one with segmentation masks and has 9,166 images with labels for two classes, with the following pixel proportion: Background  (0.95295) and Infections (0.04704). We also used a sub-set of CC-CCII with segmentation masks composed of 750 images and has labels for four classes, with the following pixel proportion: Background (0.87152), Lung Field (0.11691), \gls*{ggo} (0.00802), and Consolidation (0.00353). 

% The first step of this evaluation was the removal of labeled images that does not have any lesion from the datasets. 
% \textbf{falar sobre como foi divido os conjuntos de treino e teste}
One of the problems pointed out in~\cite{Krinski2021} was the class imbalance due to several images with just the background class; in fact, recent work has shown that several problems suffer from class imbalance~\cite{laroca2021towards, laroca2022cross}. Therefore, to mitigate this problem, in the first step of this evaluation, we removed from the datasets images with no lesion in the ground-truth mask. In the CC-CCII dataset, 4 images were removed; in the MedSeg dataset, 457 images were removed; in the MosMed, 1,264 images were removed; in the Zenodo dataset, 547 were removed; and in the Ricord1a, no image was removed. The datasets were split into 80\% for training and 20\% for testing. Then, a five-fold cross-validation strategy further divided the training set between training and validating sets. The metrics used for evaluation were the F-score described by equation~\ref{eq:dice_loss} and \gls*{iou} described by equation~\ref{eq:jaccard_loss}.

\begin{equation}
    F-score = \frac{TruePositive}{TruePositive + \frac{FalsePositive + FalseNegative}{2}}
\label{eq:dice_loss}
\end{equation}

\begin{equation}
     IoU = \frac{intersection}{union}
\label{eq:jaccard_loss}
\end{equation}

\begin{figure}[!htb]
\centering
\subfloat[][]{
	    \includegraphics[width=0.15\textwidth]{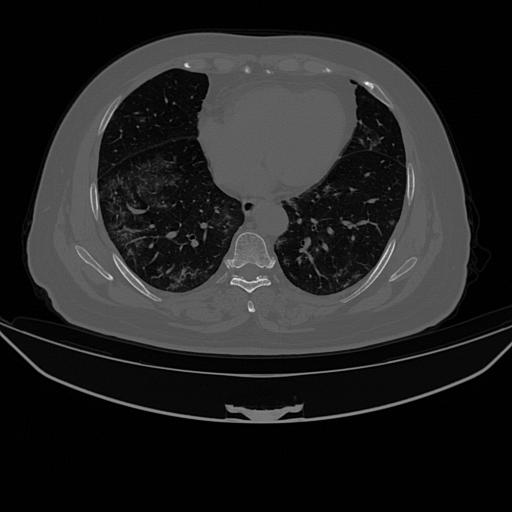}
	    \label{fig:original_image_das}
	}
	
\subfloat[][]{
	    \includegraphics[width=0.15\textwidth]{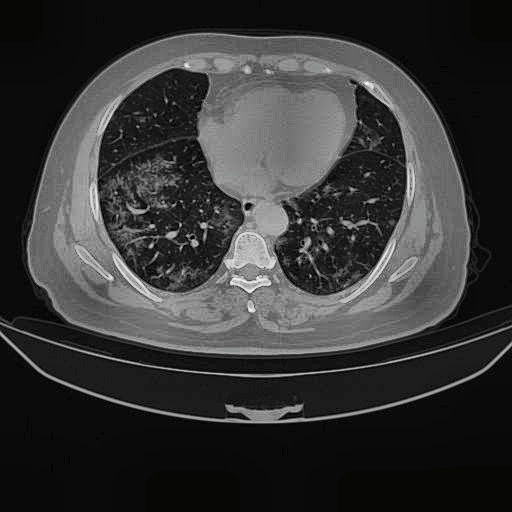}
	    \label{fig:clahe_das}
	}
\subfloat[][]{
	    \includegraphics[width=0.15\textwidth]{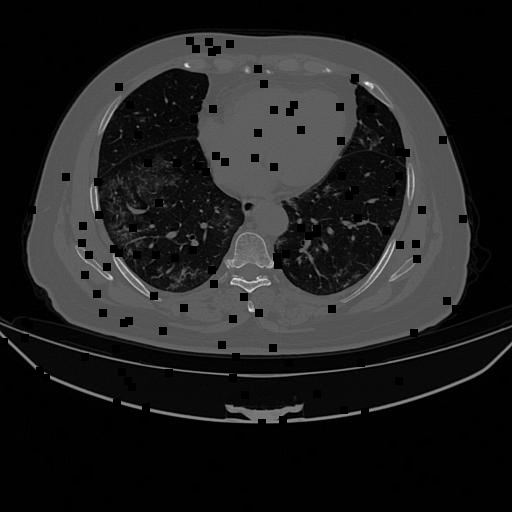}
	    \label{fig:coarse_dropout_das}
	}
\subfloat[][]{
	    \includegraphics[width=0.15\textwidth]{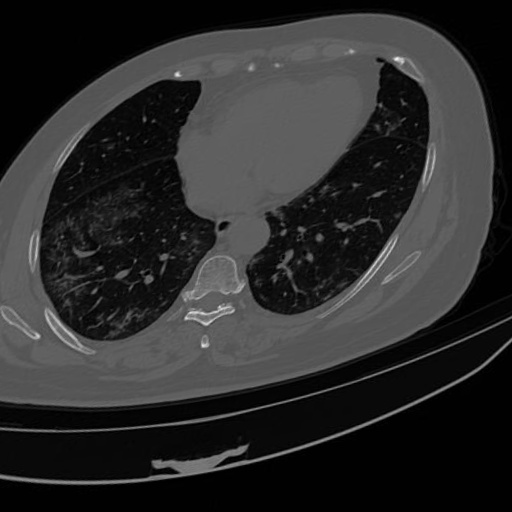}
	    \label{fig:elastic_transform_das}
	}
\subfloat[][]{
	    \includegraphics[width=0.15\textwidth]{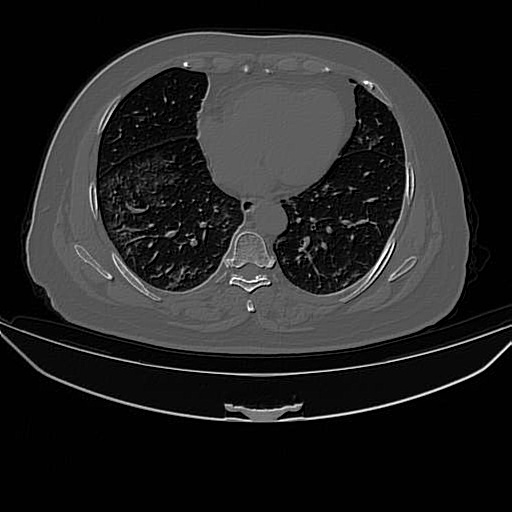}
	    \label{fig:emboss_das}
	}
\subfloat[][]{
	    \includegraphics[width=0.15\textwidth]{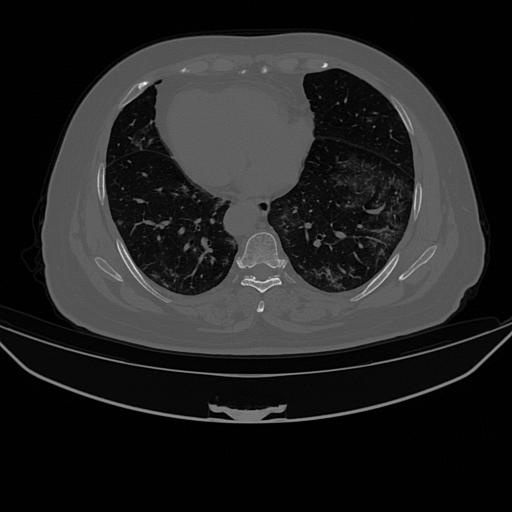}
	    \label{fig:flip_das}
	}
	
\subfloat[][]{
	    \includegraphics[width=0.15\textwidth]{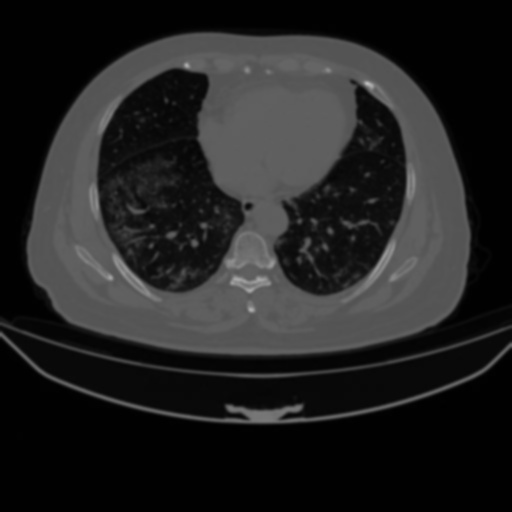}
	    \label{fig:gaussian_blur_das}
	}
\subfloat[][]{
	    \includegraphics[width=0.15\textwidth]{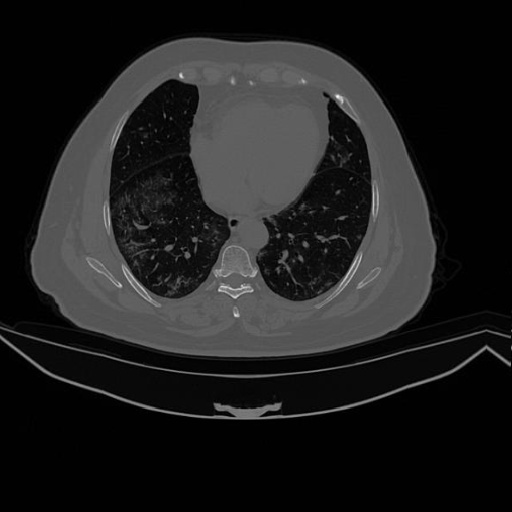}
	    \label{fig:grid_distortion_das}
	}
\subfloat[][]{
	    \includegraphics[width=0.15\textwidth]{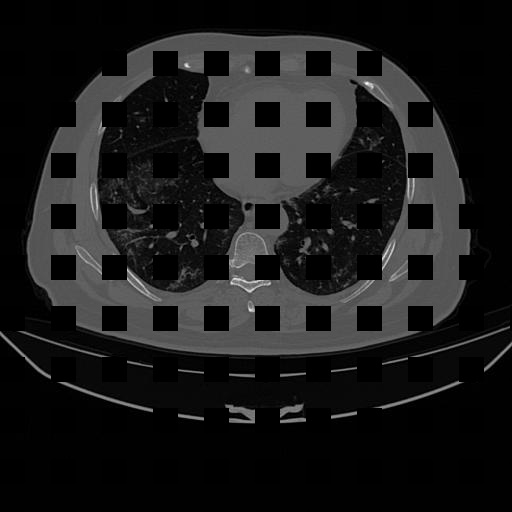}
	    \label{fig:grid_dropout_das}
	}
\subfloat[][]{
	    \includegraphics[width=0.15\textwidth]{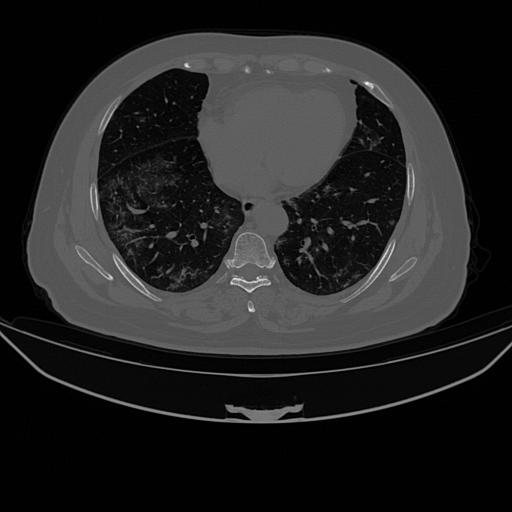}
	    \label{fig:image_Compression_das}
	}
\subfloat[][]{
	    \includegraphics[width=0.15\textwidth]{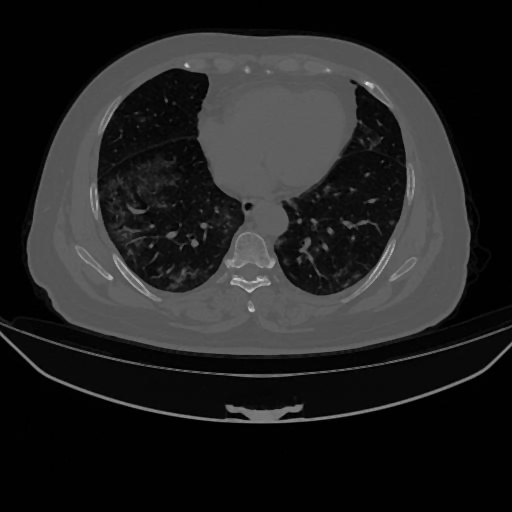}
	    \label{fig:median_blur_das}
	}
	
\subfloat[][]{
	    \includegraphics[width=0.15\textwidth]{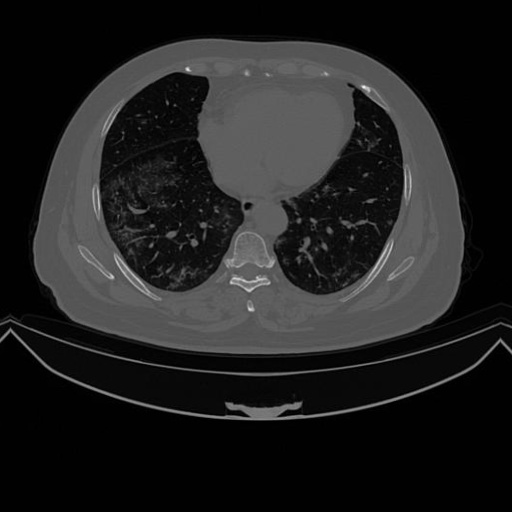}
	    \label{fig:optical_distortion_das}
	}
\subfloat[][]{
	    \includegraphics[width=0.15\textwidth]{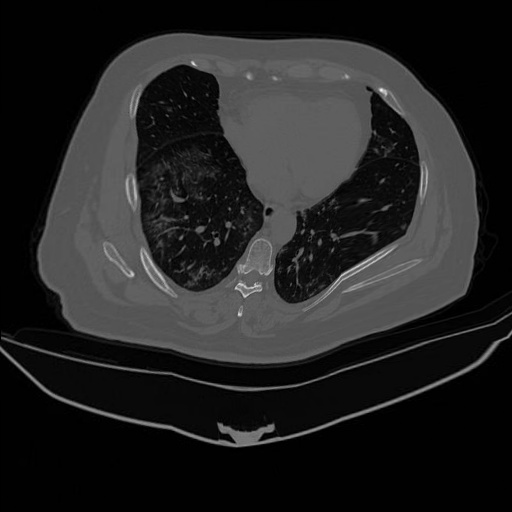}
	    \label{fig:piecewise_affine_das}
	}
\subfloat[][]{
	    \includegraphics[width=0.15\textwidth]{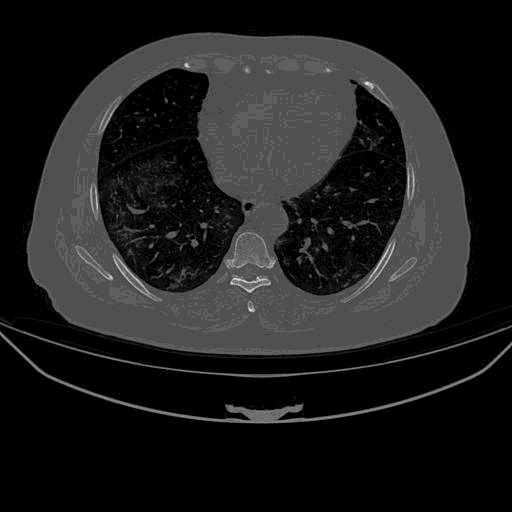}
	    \label{fig:posterize_das}
	}
\subfloat[][]{
	    \includegraphics[width=0.15\textwidth]{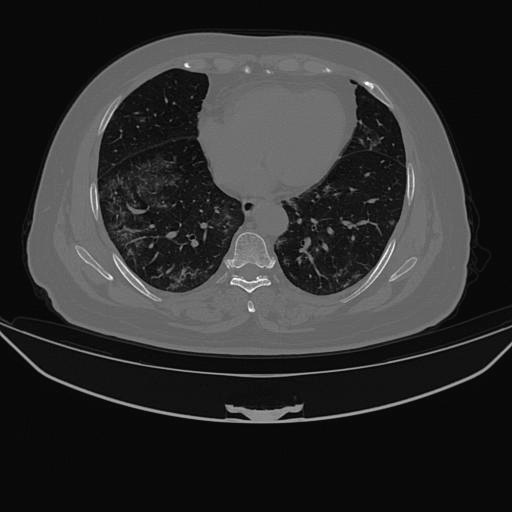}
	    \label{fig:random_brightness_contrast_das}
	}
\subfloat[][]{
	    \includegraphics[width=0.15\textwidth]{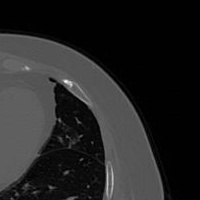}
	    \label{fig:random_crop_das}
	}
	
\subfloat[][]{
	    \includegraphics[width=0.15\textwidth]{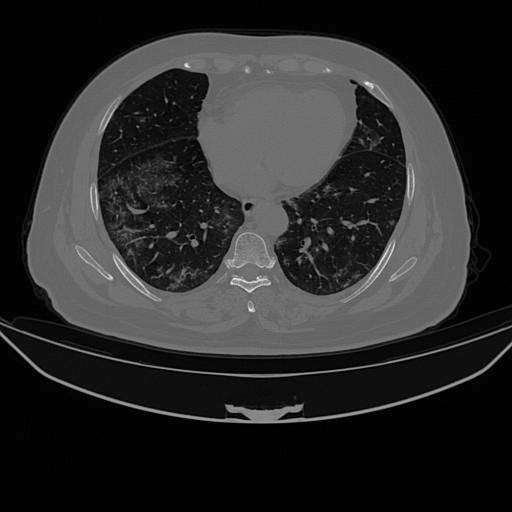}
	    \label{fig:random_gamma_das}
	}
\subfloat[][]{
	    \includegraphics[width=0.15\textwidth]{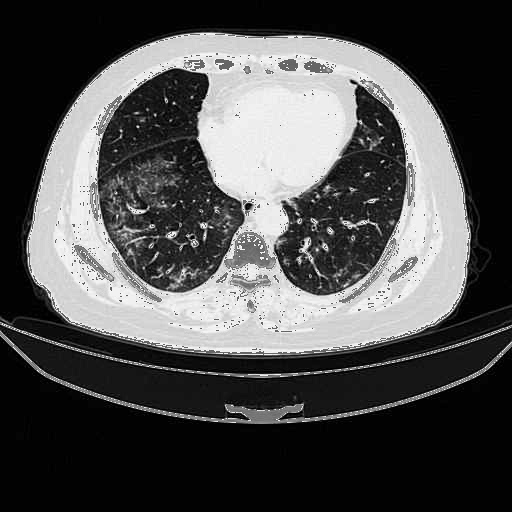}
	    \label{fig:random_snow_das}
	}
\subfloat[][]{
	    \includegraphics[width=0.15\textwidth]{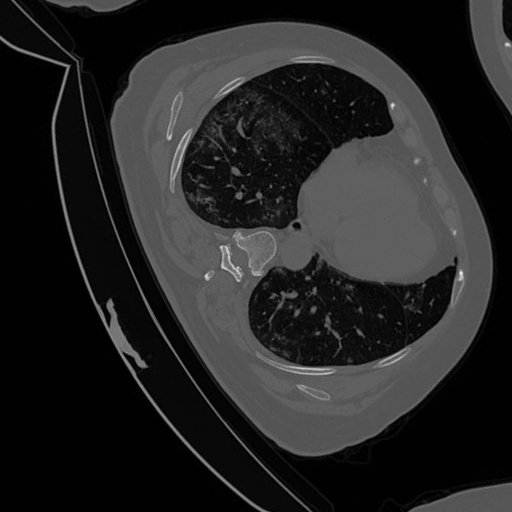}
	    \label{fig:rotate_das}
	}
\subfloat[][]{
	    \includegraphics[width=0.15\textwidth]{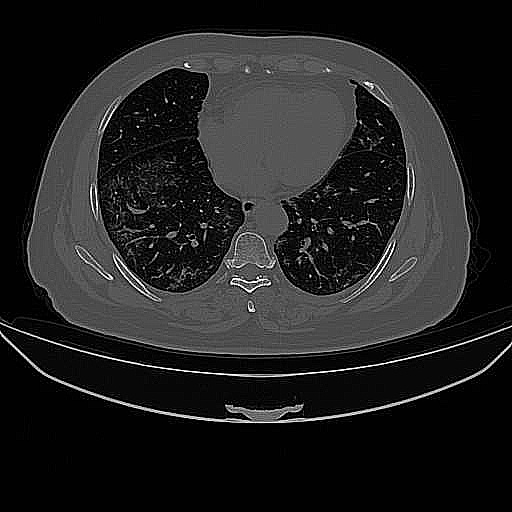}
	    \label{fig:sharpen_das}
	}
\subfloat[][]{
	    \includegraphics[width=0.15\textwidth]{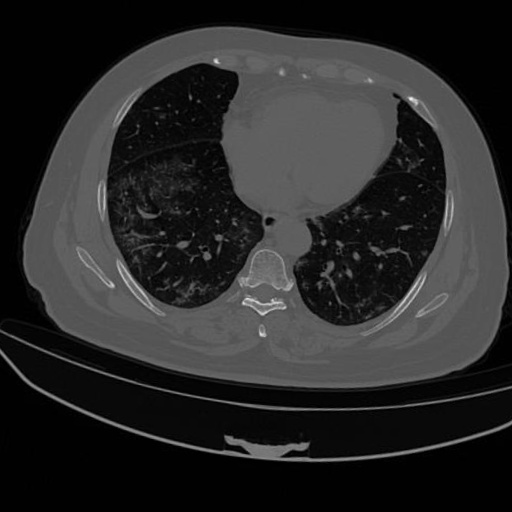}
	    \label{fig:shift_scale_rotate_das}
	}

\caption[Illustration of twenty data augmentation techniques applied in a \gls*{ct} image.]{Illustration of twenty data augmentation techniques applied in a \gls*{ct} image. In~\ref{fig:original_image_das}, the original image is presented. %In~\ref{fig:clahe_das} to~\ref{fig:shift_scale_rotate_das}, the twenty data augmentation techniques applied on top of~\ref{fig:original_image_das}.
The data augmentations are illustrated in the following sequence: \gls{clahe}~\ref{fig:clahe_das}, Coarse Dropout~\ref{fig:coarse_dropout_das}, Elastic Transform~\ref{fig:elastic_transform_das}, Emboss~\ref{fig:emboss_das}, Flip~\ref{fig:flip_das}, Gaussian Blur~\ref{fig:gaussian_blur_das}, Grid Distortion~\ref{fig:grid_distortion_das}, Grid Dropout~\ref{fig:grid_dropout_das}, Image Compression~\ref{fig:image_Compression_das}, Median Blur~\ref{fig:median_blur_das}, Optical Distortion~\ref{fig:optical_distortion_das}, Piecewise Affine Transformation~\ref{fig:piecewise_affine_das}, Posterize~\ref{fig:posterize_das}, \gls{rbc}~\ref{fig:random_brightness_contrast_das}, Random Crop~\ref{fig:random_crop_das}, Random Gamma~\ref{fig:random_gamma_das}, Random Snow~\ref{fig:random_snow_das}, Rotate~\ref{fig:rotate_das}, Sharpen~\ref{fig:sharpen_das}, Shift Scale Rotate~\ref{fig:shift_scale_rotate_das}. 
}
\label{fig:das}
\end{figure}

%%%%%%%%%%%%%%%%%%%%%%%%%%%%%%%%%%%%%%%%%%%%%%%%%%%%%%%%%%%%%%%%%%%%%%%%%%%%%%%%
\section{Experiments}
\label{sec:exp}

% In general, data augmentation~\ techniques are not evaluated in studies proposed in the COVID-19 \gls*{ct} segmentation problem, with just a few works making use of them~\cite{zhao2021d2a,qiblawey2021detection,JosephRaj2021,muller2020automated,chen2020residual,xu2020gasnet}. Also, the data augmentation techniques applied to this problem are limited to Flip and Rotation operations. To properly measure the impact of data augmentation on the  COVID-19 \gls*{ct} segmentation problem, we evaluated twenty data augmentation techniques. 

In general, when dealing with the COVID-19 \gls*{ct} segmentation problem it is usual to either completely neglect a dedicated evaluation of the impact of \gls*{da} techniques or merely report using a limited set of generic \gls*{da}, not optimized or specially designed for medical images, like Flip and Rotation operations, such as in \cite{zhao2021d2a,qiblawey2021detection,JosephRaj2021,muller2020automated,chen2020residual,xu2020gasnet}. To properly measure the impact of data augmentation on the  COVID-19 \gls*{ct} segmentation problem, we evaluated twenty data augmentation techniques. 

In this work we evaluated the following twenty data augmentation techniques: \gls{clahe}, Coarse Dropout, Elastic Transform, Emboss, Flip, Gaussian Blur, Grid Distortion, Grid Dropout, Image Compression, Median Blur, Optical Distortion, Piecewise Affine Transformation, Posterize, \gls{rbc}, Random Crop, Random Gamma, Random Snow, Rotate, Sharpen, Shift Scale Rotate. Figure~\ref{fig:das} illustrates the twenty data augmentation techniques applied to a \gls*{ct} image~\ref{fig:original_image_das}. All data augmentation techniques evaluated here are available in the Albumentation library~\cite{info11020125}.

The encoder-decoder network chosen to evaluate the dataset augmentations was the RegNetx-002~\cite{xu2021regnet} encoder and U-net++~\cite{Zhou2018} decoder. Since the encoders achieved close results in the comparison performed in~\cite{Krinski2021}, the RegNetx-002 was chosen due to being the network with a smaller number of parameters, making the RegNetx-002 faster for training. The U-net++ was chosen because it achieved the highest F-score compared with other decoders~\cite{Krinski2021}. The evaluation of how data augmentation affects the results of different encoders and decoders was left for future evaluation. Also, all experiments were evaluated through a five-fold cross-validation strategy. In total, we performed three sets of data augmentation evaluation, with each set varying network training parameters. The varied parameters were the number of epochs trained, the learning rate, and learning rate decay.

\begin{table}[!ht]
\centering
\caption[Results of the first data augmentation evaluation.]{Results of the first data augmentation evaluation in the test set. The blue-colored values indicate the best F-scores, and the red-colored values indicate the best \glspl*{iou} values. Two probabilities of applying the data augmentation were evaluated: 0.1 and 0.2. The green highlighted values show the data augmentations in which the P-value achieved values lower than 0.05, and the null hypothesis was rejected.}
\resizebox{\textwidth}{!}{%
\tiny{
\begin{tabular}{cccccccccccc}

\toprule
\multicolumn{1}{c}{\textbf{Probability}} &
\multicolumn{1}{c}{\textbf{Augmentation}} &
\multicolumn{2}{c}{\textbf{CC-CCII}} &
\multicolumn{2}{c}{\textbf{MedSeg}} &
\multicolumn{2}{c}{\textbf{MosMed}} &
\multicolumn{2}{c}{\textbf{Ricord1a}} &
\multicolumn{2}{c}{\textbf{Zenodo}} 
\\ \midrule

\makecell{} &
\makecell{} &
\makecell{F-score} &
\makecell{\gls*{iou}} & 
\makecell{F-score} &
\makecell{\gls*{iou}} & 
\makecell{F-score} &
\makecell{\gls*{iou}} & 
\makecell{F-score} &
\makecell{\gls*{iou}} & 
\makecell{F-score} &
\makecell{\gls*{iou}}
\\ \midrule

\makecell{} &
\makecell{No Augmentation}&
\makecell{0.6419}&
\makecell{0.6011}&
\makecell{0.4763}&
\makecell{0.4289}&
\makecell{0.7859}&
\makecell{0.7211}&
\makecell{0.8956}&
\makecell{0.8372}&
\makecell{0.8497}&
\makecell{0.8152}
\\ \midrule

\makecell{0.1} &
\makecell{\colorbox{white}{\makebox(10,1){CLAHE}}\\
\colorbox{white}{\makebox(10,1){Coarse Dropout}}\\
\colorbox{white}{\makebox(10,1){Elastic Transform}}\\
\colorbox{white}{\makebox(10,1){Emboss}}\\
\colorbox{white}{\makebox(10,1){Flip}}\\
\colorbox{white}{\makebox(10,1){Gaussian Blur}}\\
\colorbox{white}{\makebox(10,1){Grid Distortion}}\\
\colorbox{white}{\makebox(10,1){Grid Dropout}}\\
\colorbox{white}{\makebox(10,1){Image Compression}}\\
\colorbox{white}{\makebox(10,1){Median Blur}}\\
\colorbox{white}{\makebox(10,1){Optical Distortion}}\\
\colorbox{white}{\makebox(10,1){Piecewise Affine}}\\
\colorbox{white}{\makebox(10,1){Posterize}}\\
\colorbox{white}{\makebox(10,1){\gls{rbc}}}\\
\colorbox{white}{\makebox(10,1){Random Crop}}\\
\colorbox{white}{\makebox(10,1){Random Gamma}}\\
\colorbox{white}{\makebox(10,1){Random Snow}}\\
\colorbox{white}{\makebox(10,1){Rotate}}\\
\colorbox{white}{\makebox(10,1){Sharpen}}\\
\colorbox{white}{\makebox(10,1){Shift Scale Rotate}}\\}&
\makecell{\colorbox{white}{\makebox(10,1){0.6363}}\\\colorbox{white}{\makebox(10,1){0.6410}}\\\colorbox{green!30}{\makebox(10,1){0.6454}}\\\colorbox{white}{\makebox(10,1){0.6409}}\\\colorbox{green!30}{\makebox(10,1){0.6458}}\\\colorbox{white}{\makebox(10,1){0.6419}}\\\colorbox{green!30}{\makebox(10,1){0.6443}}\\\colorbox{white}{\makebox(10,1){0.6409}}\\\colorbox{white}{\makebox(10,1){0.6427}}\\\colorbox{white}{\makebox(10,1){0.6423}}\\\colorbox{white}{\makebox(10,1){0.6439}}\\\colorbox{white}{\makebox(10,1){0.6430}}\\\colorbox{green!30}{\makebox(10,1){0.6446}}\\\colorbox{green!30}{\makebox(10,1){0.6452}}\\\colorbox{white}{\makebox(10,1){0.6425}}\\\colorbox{white}{\makebox(10,1){0.6421}}\\\colorbox{white}{\makebox(10,1){0.6415}}\\\colorbox{green!30}{\makebox(10,1){\textbf{\textcolor{blue}{0.6473}}}}\\\colorbox{white}{\makebox(10,1){0.6426}}\\\colorbox{green!30}{\makebox(10,1){0.6452}}}&
\makecell{\colorbox{white}{\makebox(10,1){0.5962}}\\\colorbox{white}{\makebox(10,1){0.6004}}\\\colorbox{white}{\makebox(10,1){0.6048}}\\\colorbox{white}{\makebox(10,1){0.6007}}\\\colorbox{white}{\makebox(10,1){0.6044}}\\\colorbox{white}{\makebox(10,1){0.6011}}\\\colorbox{white}{\makebox(10,1){0.6036}}\\\colorbox{white}{\makebox(10,1){0.5999}}\\\colorbox{white}{\makebox(10,1){0.6019}}\\\colorbox{white}{\makebox(10,1){0.6015}}\\\colorbox{white}{\makebox(10,1){0.6032}}\\\colorbox{white}{\makebox(10,1){0.6025}}\\\colorbox{white}{\makebox(10,1){0.6039}}\\\colorbox{white}{\makebox(10,1){0.6039}}\\\colorbox{white}{\makebox(10,1){0.6020}}\\\colorbox{white}{\makebox(10,1){0.6017}}\\\colorbox{white}{\makebox(10,1){0.6010}}\\\textbf{\textcolor{red}{0.6064}}\\\colorbox{white}{\makebox(10,1){0.6024}}\\\colorbox{white}{\makebox(10,1){0.6043}}}&
\makecell{0.4754\\\colorbox{white}{\makebox(10,1){0.4756}}\\\colorbox{white}{\makebox(10,1){0.4781}}\\\colorbox{white}{\makebox(10,1){0.4749}}\\\colorbox{green!30}{\makebox(10,1){\textbf{\textcolor{blue}{0.4787}}}}\\\colorbox{white}{\makebox(10,1){0.4758}}\\\colorbox{white}{\makebox(10,1){0.4764}}\\\colorbox{white}{\makebox(10,1){0.4748}}\\\colorbox{white}{\makebox(10,1){0.4769}}\\\colorbox{white}{\makebox(10,1){0.4764}}\\\colorbox{white}{\makebox(10,1){0.4768}}\\\colorbox{white}{\makebox(10,1){0.4756}}\\\colorbox{white}{\makebox(10,1){0.4763}}\\\colorbox{white}{\makebox(10,1){0.4780}}\\\colorbox{white}{\makebox(10,1){0.4757}}\\\colorbox{white}{\makebox(10,1){0.4759}}\\\colorbox{white}{\makebox(10,1){0.4737}}\\\colorbox{white}{\makebox(10,1){0.4775}}\\\colorbox{white}{\makebox(10,1){0.4760}}\\\colorbox{white}{\makebox(10,1){0.4785}}}&
\makecell{\colorbox{white}{\makebox(10,1){0.4278}}\\\colorbox{white}{\makebox(10,1){0.4281}}\\\colorbox{white}{\makebox(10,1){0.4305}}\\\colorbox{white}{\makebox(10,1){0.4275}}\\\textbf{\textcolor{red}{0.4309}}\\\colorbox{white}{\makebox(10,1){0.4283}}\\\colorbox{white}{\makebox(10,1){0.4291}}\\\colorbox{white}{\makebox(10,1){0.4272}}\\\colorbox{white}{\makebox(10,1){0.4295}}\\\colorbox{white}{\makebox(10,1){0.4290}}\\\colorbox{white}{\makebox(10,1){0.4292}}\\\colorbox{white}{\makebox(10,1){0.4282}}\\\colorbox{white}{\makebox(10,1){0.4289}}\\\colorbox{white}{\makebox(10,1){0.4303}}\\\colorbox{white}{\makebox(10,1){0.4285}}\\\colorbox{white}{\makebox(10,1){0.4280}}\\\colorbox{white}{\makebox(10,1){0.4263}}\\\colorbox{white}{\makebox(10,1){0.4299}}\\\colorbox{white}{\makebox(10,1){0.4285}}\\\colorbox{white}{\makebox(10,1){0.4307}}}&
\makecell{\colorbox{green!30}{\makebox(10,1){0.7927}}\\\colorbox{white}{\makebox(10,1){0.7904}}\\\colorbox{green!30}{\makebox(10,1){0.7977}}\\\colorbox{green!30}{\makebox(10,1){0.7922}}\\\colorbox{green!30}{\makebox(10,1){0.7957}}\\\colorbox{white}{\makebox(10,1){0.7892}}\\\colorbox{green!30}{\makebox(10,1){0.7967}}\\\colorbox{green!30}{\makebox(10,1){0.7916}}\\\colorbox{green!30}{\makebox(10,1){0.7936}}\\\colorbox{white}{\makebox(10,1){0.7851}}\\\colorbox{green!30}{\makebox(10,1){0.7932}}\\\colorbox{green!30}{\makebox(10,1){0.8000}}\\\colorbox{green!30}{\makebox(10,1){0.7954}}\\\colorbox{white}{\makebox(10,1){0.7859}}\\\colorbox{white}{\makebox(10,1){0.7863}}\\\colorbox{green!30}{\makebox(10,1){0.7907}}\\\colorbox{white}{\makebox(10,1){0.7887}}\\\colorbox{green!30}{\makebox(10,1){0.7998}}\\\colorbox{white}{\makebox(10,1){0.7885}}\\\colorbox{green!30}{\makebox(10,1){\textbf{\textcolor{blue}{0.8001}}}}}&
\makecell{0.7268\\\colorbox{white}{\makebox(10,1){0.7244}}\\\colorbox{white}{\makebox(10,1){0.7311}}\\\colorbox{white}{\makebox(10,1){0.7265}}\\\colorbox{white}{\makebox(10,1){0.7296}}\\\colorbox{white}{\makebox(10,1){0.7238}}\\\colorbox{white}{\makebox(10,1){0.7308}}\\\colorbox{white}{\makebox(10,1){0.7265}}\\\colorbox{white}{\makebox(10,1){0.7276}}\\\colorbox{white}{\makebox(10,1){0.7201}}\\\colorbox{white}{\makebox(10,1){0.7269}}\\\colorbox{white}{\makebox(10,1){0.7335}}\\\colorbox{white}{\makebox(10,1){0.7285}}\\\colorbox{white}{\makebox(10,1){0.7202}}\\\colorbox{white}{\makebox(10,1){0.7218}}\\\colorbox{white}{\makebox(10,1){0.7252}}\\\colorbox{white}{\makebox(10,1){0.7231}}\\\colorbox{white}{\makebox(10,1){0.7335}}\\\colorbox{white}{\makebox(10,1){0.7225}}\\\textbf{\textcolor{red}{0.7336}}}&
\makecell{\colorbox{white}{\makebox(10,1){0.8933}}\\\colorbox{white}{\makebox(10,1){0.8941}}\\\colorbox{white}{\makebox(10,1){0.8915}}\\\textbf{\textcolor{blue}{0.8960}}\\\colorbox{white}{\makebox(10,1){0.8858}}\\\colorbox{white}{\makebox(10,1){0.8942}}\\\colorbox{white}{\makebox(10,1){0.8911}}\\\colorbox{white}{\makebox(10,1){0.8954}}\\\colorbox{white}{\makebox(10,1){0.8950}}\\\colorbox{white}{\makebox(10,1){0.8955}}\\\colorbox{white}{\makebox(10,1){0.8922}}\\\colorbox{white}{\makebox(10,1){0.8909}}\\\colorbox{white}{\makebox(10,1){0.8931}}\\\colorbox{white}{\makebox(10,1){0.8930}}\\\colorbox{white}{\makebox(10,1){0.8952}}\\\colorbox{white}{\makebox(10,1){0.8933}}\\\colorbox{white}{\makebox(10,1){0.8936}}\\\colorbox{white}{\makebox(10,1){0.8893}}\\\colorbox{white}{\makebox(10,1){0.8941}}\\\colorbox{white}{\makebox(10,1){0.8924}}}&
\makecell{\colorbox{white}{\makebox(10,1){0.8340}}\\\colorbox{white}{\makebox(10,1){0.8350}}\\\colorbox{white}{\makebox(10,1){0.8322}}\\\textbf{\textcolor{red}{0.8374}}\\\colorbox{white}{\makebox(10,1){0.8256}}\\\colorbox{white}{\makebox(10,1){0.8353}}\\\colorbox{white}{\makebox(10,1){0.8314}}\\\colorbox{white}{\makebox(10,1){0.8367}}\\\colorbox{white}{\makebox(10,1){0.8361}}\\\colorbox{white}{\makebox(10,1){0.8371}}\\\colorbox{white}{\makebox(10,1){0.8330}}\\\colorbox{white}{\makebox(10,1){0.8320}}\\\colorbox{white}{\makebox(10,1){0.8338}}\\\colorbox{white}{\makebox(10,1){0.8336}}\\\colorbox{white}{\makebox(10,1){0.8367}}\\\colorbox{white}{\makebox(10,1){0.8345}}\\\colorbox{white}{\makebox(10,1){0.8346}}\\\colorbox{white}{\makebox(10,1){0.8297}}\\\colorbox{white}{\makebox(10,1){0.8350}}\\\colorbox{white}{\makebox(10,1){0.8330}}}&
\makecell{\colorbox{white}{\makebox(10,1){0.8490}}\\\colorbox{white}{\makebox(10,1){0.8488}}\\\colorbox{white}{\makebox(10,1){0.8496}}\\\colorbox{white}{\makebox(10,1){0.8490}}\\\colorbox{white}{\makebox(10,1){0.8468}}\\\colorbox{white}{\makebox(10,1){0.8494}}\\\textbf{\textcolor{blue}{0.8502}}\\\colorbox{white}{\makebox(10,1){0.8493}}\\\colorbox{white}{\makebox(10,1){0.8499}}\\\colorbox{white}{\makebox(10,1){0.8496}}\\\colorbox{white}{\makebox(10,1){0.8500}}\\\colorbox{white}{\makebox(10,1){0.8499}}\\\colorbox{white}{\makebox(10,1){0.8492}}\\\colorbox{white}{\makebox(10,1){0.8493}}\\\colorbox{white}{\makebox(10,1){0.8497}}\\\colorbox{white}{\makebox(10,1){0.8496}}\\\colorbox{white}{\makebox(10,1){0.8492}}\\\colorbox{white}{\makebox(10,1){0.8475}}\\\colorbox{white}{\makebox(10,1){0.8494}}\\\colorbox{white}{\makebox(10,1){0.8494}}}&
\makecell{0.8142\\\colorbox{white}{\makebox(10,1){0.8140}}\\\colorbox{white}{\makebox(10,1){0.8149}}\\\colorbox{white}{\makebox(10,1){0.8143}}\\\colorbox{white}{\makebox(10,1){0.8110}}\\\colorbox{white}{\makebox(10,1){0.8148}}\\\textbf{\textcolor{red}{0.8156}}\\\colorbox{white}{\makebox(10,1){0.8146}}\\\colorbox{white}{\makebox(10,1){0.8155}}\\\colorbox{white}{\makebox(10,1){0.8151}}\\\colorbox{white}{\makebox(10,1){0.8155}}\\\colorbox{white}{\makebox(10,1){0.8154}}\\\colorbox{white}{\makebox(10,1){0.8147}}\\\colorbox{white}{\makebox(10,1){0.8146}}\\\colorbox{white}{\makebox(10,1){0.8151}}\\\colorbox{white}{\makebox(10,1){0.8149}}\\\colorbox{white}{\makebox(10,1){0.8143}}\\\colorbox{white}{\makebox(10,1){0.8118}}\\\colorbox{white}{\makebox(10,1){0.8149}}\\\colorbox{white}{\makebox(10,1){0.8146}}}
\\ \midrule
\makecell{0.2} &
\makecell{\colorbox{white}{\makebox(10,1){CLAHE}}\\
\colorbox{white}{\makebox(10,1){Coarse Dropout}}\\
\colorbox{white}{\makebox(10,1){Elastic Transform}}\\
\colorbox{white}{\makebox(10,1){Emboss}}\\
\colorbox{white}{\makebox(10,1){Flip}}\\
\colorbox{white}{\makebox(10,1){Gaussian Blur}}\\
\colorbox{white}{\makebox(10,1){Grid Distortion}}\\
\colorbox{white}{\makebox(10,1){Grid Dropout}}\\
\colorbox{white}{\makebox(10,1){Image Compression}}\\
\colorbox{white}{\makebox(10,1){Median Blur}}\\
\colorbox{white}{\makebox(10,1){Optical Distortion}}\\
\colorbox{white}{\makebox(10,1){Piecewise Affine}}\\
\colorbox{white}{\makebox(10,1){Posterize}}\\
\colorbox{white}{\makebox(10,1){\gls{rbc}}}\\
\colorbox{white}{\makebox(10,1){Random Crop}}\\
\colorbox{white}{\makebox(10,1){Random Gamma}}\\
\colorbox{white}{\makebox(10,1){Random Snow}}\\
\colorbox{white}{\makebox(10,1){Rotate}}\\
\colorbox{white}{\makebox(10,1){Sharpen}}\\
\colorbox{white}{\makebox(10,1){Shift Scale Rotate}}\\}&
\makecell{\colorbox{white}{\makebox(10,1){0.6386}}\\\colorbox{white}{\makebox(10,1){0.6379}}\\\colorbox{green!30}{\makebox(10,1){0.6454}}\\\colorbox{white}{\makebox(10,1){0.6420}}\\\colorbox{green!30}{\makebox(10,1){0.6465}}\\\colorbox{white}{\makebox(10,1){0.6421}}\\\colorbox{green!30}{\makebox(10,1){0.6476}}\\\colorbox{white}{\makebox(10,1){0.6398}}\\\colorbox{white}{\makebox(10,1){0.6436}}\\\colorbox{green!30}{\makebox(10,1){0.6431}}\\\colorbox{green!30}{\makebox(10,1){0.6455}}\\\colorbox{green!30}{\makebox(10,1){0.6480}}\\\colorbox{white}{\makebox(10,1){0.6417}}\\\colorbox{white}{\makebox(10,1){0.6419}}\\\colorbox{green!30}{\makebox(10,1){0.6456}}\\\colorbox{white}{\makebox(10,1){0.6415}}\\\colorbox{white}{\makebox(10,1){0.6422}}\\\colorbox{white}{\makebox(10,1){0.6458}}\\\colorbox{white}{\makebox(10,1){0.6418}}\\\colorbox{green!30}{\makebox(10,1){\textbf{\textcolor{blue}{0.6482}}}}}&
\makecell{\colorbox{white}{\makebox(10,1){0.5978}}\\\colorbox{white}{\makebox(10,1){0.5976}}\\\colorbox{white}{\makebox(10,1){0.6050}}\\\colorbox{white}{\makebox(10,1){0.6015}}\\\colorbox{white}{\makebox(10,1){0.6054}}\\\colorbox{white}{\makebox(10,1){0.6015}}\\\colorbox{white}{\makebox(10,1){0.6070}}\\\colorbox{white}{\makebox(10,1){0.5990}}\\\colorbox{white}{\makebox(10,1){0.6022}}\\\colorbox{white}{\makebox(10,1){0.6025}}\\\colorbox{white}{\makebox(10,1){0.6045}}\\\textbf{\textcolor{red}{0.6072}}\\\colorbox{white}{\makebox(10,1){0.6014}}\\\colorbox{white}{\makebox(10,1){0.6009}}\\\colorbox{white}{\makebox(10,1){0.6044}}\\\colorbox{white}{\makebox(10,1){0.6006}}\\\colorbox{white}{\makebox(10,1){0.6015}}\\\colorbox{white}{\makebox(10,1){0.6050}}\\\colorbox{white}{\makebox(10,1){0.6009}}\\\colorbox{white}{\makebox(10,1){0.6070}}}&
\makecell{\colorbox{white}{\makebox(10,1){0.4723}}\\\colorbox{white}{\makebox(10,1){0.4744}}\\\colorbox{white}{\makebox(10,1){0.4785}}\\\colorbox{white}{\makebox(10,1){0.4755}}\\\colorbox{white}{\makebox(10,1){0.4771}}\\\colorbox{white}{\makebox(10,1){0.4755}}\\\colorbox{green!30}{\makebox(10,1){0.4802}}\\\colorbox{white}{\makebox(10,1){0.4740}}\\\colorbox{white}{\makebox(10,1){0.4776}}\\\colorbox{white}{\makebox(10,1){0.4762}}\\\colorbox{white}{\makebox(10,1){0.4781}}\\\colorbox{white}{\makebox(10,1){0.4771}}\\\colorbox{white}{\makebox(10,1){0.4772}}\\\colorbox{white}{\makebox(10,1){0.4761}}\\\colorbox{white}{\makebox(10,1){0.4752}}\\\colorbox{white}{\makebox(10,1){0.4761}}\\\colorbox{white}{\makebox(10,1){0.4711}}\\\colorbox{white}{\makebox(10,1){0.4781}}\\\colorbox{white}{\makebox(10,1){0.4770}}\\\colorbox{green!30}{\makebox(10,1){\textbf{\textcolor{blue}{0.4806}}}}}&
\makecell{\colorbox{white}{\makebox(10,1){0.4252}}\\\colorbox{white}{\makebox(10,1){0.4270}}\\\colorbox{white}{\makebox(10,1){0.4303}}\\\colorbox{white}{\makebox(10,1){0.4280}}\\\colorbox{white}{\makebox(10,1){0.4292}}\\\colorbox{white}{\makebox(10,1){0.4288}}\\\colorbox{white}{\makebox(10,1){0.4325}}\\\colorbox{white}{\makebox(10,1){0.4264}}\\\colorbox{white}{\makebox(10,1){0.4299}}\\\colorbox{white}{\makebox(10,1){0.4289}}\\\colorbox{white}{\makebox(10,1){0.4300}}\\\colorbox{white}{\makebox(10,1){0.4297}}\\\colorbox{white}{\makebox(10,1){0.4299}}\\\colorbox{white}{\makebox(10,1){0.4280}}\\\colorbox{white}{\makebox(10,1){0.4279}}\\\colorbox{white}{\makebox(10,1){0.4279}}\\\colorbox{white}{\makebox(10,1){0.4237}}\\\colorbox{white}{\makebox(10,1){0.4302}}\\\colorbox{white}{\makebox(10,1){0.4289}}\\\textbf{\textcolor{red}{0.4326}}}&
\makecell{\colorbox{white}{\makebox(10,1){0.7844}}\\\colorbox{green!30}{\makebox(10,1){0.7931}}\\\colorbox{green!30}{\makebox(10,1){0.8015}}\\\colorbox{green!30}{\makebox(10,1){0.7927}}\\\colorbox{green!30}{\makebox(10,1){0.7991}}\\\colorbox{white}{\makebox(10,1){0.7876}}\\\colorbox{green!30}{\makebox(10,1){0.8038}}\\\colorbox{white}{\makebox(10,1){0.7868}}\\\colorbox{green!30}{\makebox(10,1){0.7965}}\\\colorbox{white}{\makebox(10,1){0.7888}}\\\colorbox{green!30}{\makebox(10,1){0.7941}}\\\colorbox{green!30}{\makebox(10,1){\textbf{\textcolor{blue}{0.8049}}}}\\\colorbox{green!30}{\makebox(10,1){0.7967}}\\\colorbox{white}{\makebox(10,1){0.7875}}\\\colorbox{green!30}{\makebox(10,1){0.7943}}\\\colorbox{green!30}{\makebox(10,1){0.7935}}\\\colorbox{white}{\makebox(10,1){0.7875}}\\\colorbox{green!30}{\makebox(10,1){0.8010}}\\\colorbox{white}{\makebox(10,1){0.7852}}\\\colorbox{green!30}{\makebox(10,1){0.8004}}}&
\makecell{\colorbox{white}{\makebox(10,1){0.7192}}\\\colorbox{white}{\makebox(10,1){0.7273}}\\\colorbox{white}{\makebox(10,1){0.7345}}\\\colorbox{white}{\makebox(10,1){0.7270}}\\\colorbox{white}{\makebox(10,1){0.7328}}\\\colorbox{white}{\makebox(10,1){0.7230}}\\\colorbox{white}{\makebox(10,1){0.7368}}\\\colorbox{white}{\makebox(10,1){0.7209}}\\\colorbox{white}{\makebox(10,1){0.7310}}\\\colorbox{white}{\makebox(10,1){0.7231}}\\\colorbox{white}{\makebox(10,1){0.7279}}\\\textbf{\textcolor{red}{0.7385}}\\\colorbox{white}{\makebox(10,1){0.7317}}\\\colorbox{white}{\makebox(10,1){0.7212}}\\\colorbox{white}{\makebox(10,1){0.7287}}\\\colorbox{white}{\makebox(10,1){0.7274}}\\\colorbox{white}{\makebox(10,1){0.7215}}\\\colorbox{white}{\makebox(10,1){0.7340}}\\\colorbox{white}{\makebox(10,1){0.7189}}\\\colorbox{white}{\makebox(10,1){0.7340}}}&
\makecell{\colorbox{white}{\makebox(10,1){0.8955}}\\\colorbox{white}{\makebox(10,1){0.8920}}\\\colorbox{white}{\makebox(10,1){0.8839}}\\\colorbox{white}{\makebox(10,1){0.8942}}\\\colorbox{white}{\makebox(10,1){0.8889}}\\\colorbox{white}{\makebox(10,1){0.8928}}\\\colorbox{white}{\makebox(10,1){0.8886}}\\\colorbox{white}{\makebox(10,1){0.8855}}\\\textbf{\textcolor{blue}{0.8963}}\\\colorbox{white}{\makebox(10,1){0.8951}}\\\colorbox{white}{\makebox(10,1){0.8938}}\\\colorbox{white}{\makebox(10,1){0.8899}}\\\colorbox{white}{\makebox(10,1){0.8923}}\\\colorbox{white}{\makebox(10,1){0.8867}}\\\colorbox{white}{\makebox(10,1){0.8953}}\\\colorbox{white}{\makebox(10,1){0.8916}}\\\colorbox{white}{\makebox(10,1){0.8883}}\\\colorbox{white}{\makebox(10,1){0.8813}}\\\colorbox{white}{\makebox(10,1){0.8954}}\\\colorbox{white}{\makebox(10,1){0.8856}}}&
\makecell{\colorbox{white}{\makebox(10,1){0.8367}}\\\colorbox{white}{\makebox(10,1){0.8329}}\\\colorbox{white}{\makebox(10,1){0.8229}}\\\colorbox{white}{\makebox(10,1){0.8355}}\\\colorbox{white}{\makebox(10,1){0.8289}}\\\colorbox{white}{\makebox(10,1){0.8338}}\\\colorbox{white}{\makebox(10,1){0.8288}}\\\colorbox{white}{\makebox(10,1){0.8250}}\\\textbf{\textcolor{red}{0.8378}}\\\colorbox{white}{\makebox(10,1){0.8364}}\\\colorbox{white}{\makebox(10,1){0.8352}}\\\colorbox{white}{\makebox(10,1){0.8304}}\\\colorbox{white}{\makebox(10,1){0.8330}}\\\colorbox{white}{\makebox(10,1){0.8266}}\\\colorbox{white}{\makebox(10,1){0.8368}}\\\colorbox{white}{\makebox(10,1){0.8323}}\\\colorbox{white}{\makebox(10,1){0.8284}}\\\colorbox{white}{\makebox(10,1){0.8203}}\\\colorbox{white}{\makebox(10,1){0.8370}}\\\colorbox{white}{\makebox(10,1){0.8253}}}&
\makecell{\colorbox{white}{\makebox(10,1){0.8491}}\\\colorbox{white}{\makebox(10,1){0.8489}}\\\colorbox{white}{\makebox(10,1){0.8491}}\\\colorbox{white}{\makebox(10,1){0.8487}}\\\colorbox{white}{\makebox(10,1){0.8444}}\\\colorbox{white}{\makebox(10,1){0.8494}}\\\colorbox{white}{\makebox(10,1){0.8496}}\\\colorbox{white}{\makebox(10,1){0.8487}}\\\colorbox{white}{\makebox(10,1){0.8499}}\\\colorbox{white}{\makebox(10,1){0.8497}}\\\textbf{\textcolor{blue}{0.8503}}\\\colorbox{white}{\makebox(10,1){0.8495}}\\\colorbox{white}{\makebox(10,1){0.8497}}\\\colorbox{white}{\makebox(10,1){0.8493}}\\\colorbox{white}{\makebox(10,1){0.8497}}\\\colorbox{white}{\makebox(10,1){0.8493}}\\\colorbox{white}{\makebox(10,1){0.8484}}\\\colorbox{white}{\makebox(10,1){0.8473}}\\\colorbox{white}{\makebox(10,1){0.8495}}\\\colorbox{white}{\makebox(10,1){0.8491}}}&
\makecell{\colorbox{white}{\makebox(10,1){0.8142}}\\\colorbox{white}{\makebox(10,1){0.8141}}\\\colorbox{white}{\makebox(10,1){0.8141}}\\\colorbox{white}{\makebox(10,1){0.8137}}\\\colorbox{white}{\makebox(10,1){0.8080}}\\\colorbox{white}{\makebox(10,1){0.8146}}\\\colorbox{white}{\makebox(10,1){0.8148}}\\\colorbox{white}{\makebox(10,1){0.8140}}\\\colorbox{white}{\makebox(10,1){0.8155}}\\\colorbox{white}{\makebox(10,1){0.8150}}\\\textbf{\textcolor{red}{0.8159}}\\\colorbox{white}{\makebox(10,1){0.8147}}\\\colorbox{white}{\makebox(10,1){0.8151}}\\\colorbox{white}{\makebox(10,1){0.8145}}\\\colorbox{white}{\makebox(10,1){0.8150}}\\\colorbox{white}{\makebox(10,1){0.8146}}\\\colorbox{white}{\makebox(10,1){0.8133}}\\\colorbox{white}{\makebox(10,1){0.8114}}\\\colorbox{white}{\makebox(10,1){0.8148}}\\\colorbox{white}{\makebox(10,1){0.8141}}}

\\ \bottomrule
\end{tabular}
}
}
\label{tab:tests3}
\end{table}

In the first evaluation of data augmentations, the architecture was trained for 50 epochs with a learning rate of 0.001. The learning rate was divided by 10 every 10 epochs. Two probabilities of applying the data augmentation were evaluated: 0.1 and 0.2. As presented in Table~\ref{tab:tests3}, most of the data augmentations did not improve the F-score and the \gls*{iou}. The MosMed dataset was the only dataset that applying data augmentations improved the results, with improvements in the F-score of 1\% and 2\% in most of the data augmentations applied. In the Zenodo dataset, most of the augmentations achieved similar results with the baseline. However, the Grid Distortion with the probability of 0.1 and the Optical Distortion with the probability of 0.2 improved the F-score by 1\%. In the MedSeg, only the Shift Scale Rotate augmentation with the probability of 0.2 achieved better results with an F-score 1\% higher than the baseline. The CC-CCII and Ricord1a datasets did not achieve improvements with data augmentations. To perform a statistical analysis of the data augmentation evaluation, the one-sided Wilcoxon signed-rank test was applied with the null hypothesis as the F-scores of the distribution without data augmentation are greater than the distributions with data augmentation. %The green highlighted values in Table\ref{tab:tests3} show the data augmentations in which the P-value achieved values lower than 0.05, and the null hypothesis was rejected.

The datasets CC-CCII and MosMed were the datasets most sensitive to data augmentation and achieved better F-scores in seven data augmentations when the data augmentation was applied with probability 0.1 and eight data augmentations when applied with probability 0.2. In the MosMed dataset, the null hypothesis was rejected in most of the data augmentations applied in both probabilities. In the MedSeg, the null hypotheses were rejected in only one data augmentation when applied with probability 0.1 and two data augmentations when applied with probability 0.2. The Ricord1a and Zenodo did not achieve statistical significance to reject the null hypotheses in any data augmentation.

\begin{table}[!ht]
\centering
\caption[Results of the second data augmentation evaluation.]{Results of the second data augmentation evaluation in the test set. The blue-colored values indicate the best F-scores, and the red-colored values indicate the best \glspl*{iou} values. Two probabilities of applying the data augmentation were evaluated: 0.1 and 0.2. The green highlighted values show the data augmentations in which the P-value achieved values lower than 0.05, and the null hypothesis was rejected.} 

\resizebox{\textwidth}{!}{%
\tiny{
\begin{tabular}{cccccccccccc}

\toprule
\multicolumn{1}{c}{\textbf{Probability}} &
\multicolumn{1}{c}{\textbf{Augmentation}} &
\multicolumn{2}{c}{\textbf{CC-CCII}} &
\multicolumn{2}{c}{\textbf{MedSeg}} &
\multicolumn{2}{c}{\textbf{MosMed}} &
\multicolumn{2}{c}{\textbf{Ricord1a}} &
\multicolumn{2}{c}{\textbf{Zenodo}} 
\\ \midrule

\makecell{} &
\makecell{} &
\makecell{F-score} &
\makecell{\gls*{iou}} & 
\makecell{F-score} &
\makecell{\gls*{iou}} & 
\makecell{F-score} &
\makecell{\gls*{iou}} & 
\makecell{F-score} &
\makecell{\gls*{iou}} & 
\makecell{F-score} &
\makecell{\gls*{iou}}
\\ \midrule

\makecell{} &
\makecell{No Augmentation}&
\makecell{0.6447}&
\makecell{0.6048}&
\makecell{0.4780}&
\makecell{0.4308}&
\makecell{0.7938}&
\makecell{0.7292}&
\makecell{0.9031}&
\makecell{0.8472}&
\makecell{0.8511}&
\makecell{0.8174}
\\ \midrule

\makecell{0.1} &
\makecell{\colorbox{white}{\makebox(10,1){CLAHE}}\\
\colorbox{white}{\makebox(10,1){Coarse Dropout}}\\
\colorbox{white}{\makebox(10,1){Elastic Transform}}\\
\colorbox{white}{\makebox(10,1){Emboss}}\\
\colorbox{white}{\makebox(10,1){Flip}}\\
\colorbox{white}{\makebox(10,1){Gaussian Blur}}\\
\colorbox{white}{\makebox(10,1){Grid Distortion}}\\
\colorbox{white}{\makebox(10,1){Grid Dropout}}\\
\colorbox{white}{\makebox(10,1){Image Compression}}\\
\colorbox{white}{\makebox(10,1){Median Blur}}\\
\colorbox{white}{\makebox(10,1){Optical Distortion}}\\
\colorbox{white}{\makebox(10,1){Piecewise Affine}}\\
\colorbox{white}{\makebox(10,1){Posterize}}\\
\colorbox{white}{\makebox(10,1){\gls{rbc}}}\\
\colorbox{white}{\makebox(10,1){Random Crop}}\\
\colorbox{white}{\makebox(10,1){Random Gamma}}\\
\colorbox{white}{\makebox(10,1){Random Snow}}\\
\colorbox{white}{\makebox(10,1){Rotate}}\\
\colorbox{white}{\makebox(10,1){Sharpen}}\\
\colorbox{white}{\makebox(10,1){Shift Scale Rotate}}}&
\makecell{\colorbox{white}{\makebox(10,1){0.6431}}\\\colorbox{white}{\makebox(10,1){0.6429}}\\\colorbox{green!30}{\makebox(10,1){0.6499}}\\\colorbox{white}{\makebox(10,1){0.6458}}\\\colorbox{green!30}{\makebox(10,1){0.6493}}\\\colorbox{white}{\makebox(10,1){0.6448}}\\\colorbox{green!30}{\makebox(10,1){0.6506}}\\\colorbox{white}{\makebox(10,1){0.6412}}\\\colorbox{white}{\makebox(10,1){0.6462}}\\\colorbox{white}{\makebox(10,1){0.6462}}\\\colorbox{green!30}{\makebox(10,1){0.6487}}\\\colorbox{green!30}{\makebox(10,1){0.6495}}\\\colorbox{white}{\makebox(10,1){0.6469}}\\\colorbox{white}{\makebox(10,1){0.6439}}\\\colorbox{white}{\makebox(10,1){0.6470}}\\\colorbox{white}{\makebox(10,1){0.6449}}\\\colorbox{white}{\makebox(10,1){0.6435}}\\\colorbox{green!30}{\makebox(10,1){\textbf{\textcolor{blue}{0.6522}}}}\\\colorbox{white}{\makebox(10,1){0.6441}}\\\colorbox{green!30}{\makebox(10,1){0.6504}}}&
\makecell{\colorbox{white}{\makebox(10,1){0.6025}}\\\colorbox{white}{\makebox(10,1){0.6016}}\\\colorbox{white}{\makebox(10,1){0.6095}}\\\colorbox{white}{\makebox(10,1){0.6054}}\\\colorbox{white}{\makebox(10,1){0.6087}}\\\colorbox{white}{\makebox(10,1){0.6045}}\\\colorbox{white}{\makebox(10,1){0.6102}}\\\colorbox{white}{\makebox(10,1){0.6011}}\\\colorbox{white}{\makebox(10,1){0.6052}}\\\colorbox{white}{\makebox(10,1){0.6060}}\\\colorbox{white}{\makebox(10,1){0.6079}}\\\colorbox{white}{\makebox(10,1){0.6092}}\\\colorbox{white}{\makebox(10,1){0.6070}}\\\colorbox{white}{\makebox(10,1){0.6037}}\\\colorbox{white}{\makebox(10,1){0.6057}}\\\colorbox{white}{\makebox(10,1){0.6043}}\\\colorbox{white}{\makebox(10,1){0.6037}}\\\textbf{\textcolor{red}{0.6115}}\\\colorbox{white}{\makebox(10,1){0.6036}}\\\colorbox{white}{\makebox(10,1){0.6096}}}&
\makecell{\colorbox{white}{\makebox(10,1){0.4781}}\\\colorbox{white}{\makebox(10,1){0.4769}}\\\colorbox{green!30}{\makebox(10,1){0.4817}}\\\colorbox{white}{\makebox(10,1){0.4777}}\\\colorbox{green!30}{\makebox(10,1){0.4812}}\\\colorbox{green!30}{\makebox(10,1){0.4799}}\\\colorbox{green!30}{\makebox(10,1){0.4814}}\\\colorbox{green!30}{\makebox(10,1){0.4802}}\\\colorbox{white}{\makebox(10,1){0.4787}}\\\colorbox{white}{\makebox(10,1){0.4765}}\\\colorbox{green!30}{\makebox(10,1){0.4819}}\\\colorbox{green!30}{\makebox(10,1){0.4819}}\\\colorbox{green!30}{\makebox(10,1){0.4796}}\\\colorbox{green!30}{\makebox(10,1){0.4805}}\\\colorbox{white}{\makebox(10,1){0.4772}}\\\colorbox{white}{\makebox(10,1){0.4783}}\\\colorbox{white}{\makebox(10,1){0.4770}}\\\colorbox{green!30}{\makebox(10,1){0.4810}}\\\colorbox{white}{\makebox(10,1){0.4778}}\\\colorbox{green!30}{\makebox(10,1){\textbf{\textcolor{blue}{0.4823}}}}}&
\makecell{\colorbox{white}{\makebox(10,1){0.4305}}\\\colorbox{white}{\makebox(10,1){0.4304}}\\\colorbox{white}{\makebox(10,1){0.4342}}\\\colorbox{white}{\makebox(10,1){0.4308}}\\\colorbox{white}{\makebox(10,1){0.4344}}\\\colorbox{white}{\makebox(10,1){0.4329}}\\\colorbox{white}{\makebox(10,1){0.4341}}\\\colorbox{white}{\makebox(10,1){0.4324}}\\\colorbox{white}{\makebox(10,1){0.4315}}\\\colorbox{white}{\makebox(10,1){0.4299}}\\\colorbox{white}{\makebox(10,1){0.4341}}\\\colorbox{white}{\makebox(10,1){0.4349}}\\\colorbox{white}{\makebox(10,1){0.4326}}\\\colorbox{white}{\makebox(10,1){0.4326}}\\\colorbox{white}{\makebox(10,1){0.4303}}\\\colorbox{white}{\makebox(10,1){0.4305}}\\\colorbox{white}{\makebox(10,1){0.4301}}\\\colorbox{white}{\makebox(10,1){0.4341}}\\\colorbox{white}{\makebox(10,1){0.4306}}\\\textbf{\textcolor{red}{0.4350}}}&
\makecell{\colorbox{white}{\makebox(10,1){0.7939}}\\\colorbox{white}{\makebox(10,1){0.7985}}\\\colorbox{green!30}{\makebox(10,1){0.8065}}\\\colorbox{white}{\makebox(10,1){0.7953}}\\\colorbox{green!30}{\makebox(10,1){0.8115}}\\\colorbox{white}{\makebox(10,1){0.7917}}\\\colorbox{green!30}{\makebox(10,1){\textbf{\textcolor{blue}{0.8121}}}}\\\colorbox{white}{\makebox(10,1){0.7927}}\\\colorbox{white}{\makebox(10,1){0.7961}}\\\colorbox{white}{\makebox(10,1){0.7959}}\\\colorbox{green!30}{\makebox(10,1){0.8050}}\\\colorbox{green!30}{\makebox(10,1){0.8018}}\\\colorbox{white}{\makebox(10,1){0.7985}}\\\colorbox{white}{\makebox(10,1){0.7940}}\\\colorbox{white}{\makebox(10,1){0.7915}}\\\colorbox{white}{\makebox(10,1){0.7949}}\\\colorbox{white}{\makebox(10,1){0.7958}}\\\colorbox{green!30}{\makebox(10,1){0.8073}}\\\colorbox{white}{\makebox(10,1){0.7915}}\\\colorbox{green!30}{\makebox(10,1){0.8060}}}&
\makecell{\colorbox{white}{\makebox(10,1){0.7288}}\\\colorbox{white}{\makebox(10,1){0.7335}}\\\colorbox{white}{\makebox(10,1){0.7404}}\\\colorbox{white}{\makebox(10,1){0.7296}}\\\colorbox{white}{\makebox(10,1){0.7451}}\\\colorbox{white}{\makebox(10,1){0.7265}}\\\textbf{\textcolor{red}{0.7459}}\\\colorbox{white}{\makebox(10,1){0.7279}}\\\colorbox{white}{\makebox(10,1){0.7318}}\\\colorbox{white}{\makebox(10,1){0.7314}}\\\colorbox{white}{\makebox(10,1){0.7391}}\\\colorbox{white}{\makebox(10,1){0.7370}}\\\colorbox{white}{\makebox(10,1){0.7329}}\\\colorbox{white}{\makebox(10,1){0.7276}}\\\colorbox{white}{\makebox(10,1){0.7264}}\\\colorbox{white}{\makebox(10,1){0.7300}}\\\colorbox{white}{\makebox(10,1){0.7296}}\\\colorbox{white}{\makebox(10,1){0.7416}}\\\colorbox{white}{\makebox(10,1){0.7269}}\\\colorbox{white}{\makebox(10,1){0.7397}}}&
\makecell{\colorbox{white}{\makebox(10,1){0.9033}}\\\colorbox{white}{\makebox(10,1){0.9027}}\\\colorbox{white}{\makebox(10,1){0.8993}}\\\textbf{\textcolor{blue}{0.9039}}\\\colorbox{white}{\makebox(10,1){0.8990}}\\\colorbox{white}{\makebox(10,1){0.9035}}\\\colorbox{white}{\makebox(10,1){0.9022}}\\\colorbox{white}{\makebox(10,1){0.9019}}\\\colorbox{white}{\makebox(10,1){0.9033}}\\\colorbox{white}{\makebox(10,1){0.9023}}\\\colorbox{white}{\makebox(10,1){0.9035}}\\\colorbox{white}{\makebox(10,1){0.9022}}\\\colorbox{white}{\makebox(10,1){0.9021}}\\\colorbox{white}{\makebox(10,1){0.9025}}\\\colorbox{white}{\makebox(10,1){0.9009}}\\\colorbox{white}{\makebox(10,1){0.9020}}\\\colorbox{white}{\makebox(10,1){0.9027}}\\\colorbox{white}{\makebox(10,1){0.9007}}\\\colorbox{white}{\makebox(10,1){0.9026}}\\\colorbox{white}{\makebox(10,1){0.9026}}}&
\makecell{\colorbox{white}{\makebox(10,1){0.8476}}\\\colorbox{white}{\makebox(10,1){0.8464}}\\\colorbox{white}{\makebox(10,1){0.8421}}\\\textbf{\textcolor{red}{0.8482}}\\\colorbox{white}{\makebox(10,1){0.8417}}\\\colorbox{white}{\makebox(10,1){0.8473}}\\\colorbox{white}{\makebox(10,1){0.8452}}\\\colorbox{white}{\makebox(10,1){0.8453}}\\\colorbox{white}{\makebox(10,1){0.8472}}\\\colorbox{white}{\makebox(10,1){0.8458}}\\\colorbox{white}{\makebox(10,1){0.8473}}\\\colorbox{white}{\makebox(10,1){0.8454}}\\\colorbox{white}{\makebox(10,1){0.8457}}\\\colorbox{white}{\makebox(10,1){0.8460}}\\\colorbox{white}{\makebox(10,1){0.8441}}\\\colorbox{white}{\makebox(10,1){0.8455}}\\\colorbox{white}{\makebox(10,1){0.8463}}\\\colorbox{white}{\makebox(10,1){0.8439}}\\\colorbox{white}{\makebox(10,1){0.8467}}\\\colorbox{white}{\makebox(10,1){0.8460}}}&
\makecell{\colorbox{white}{\makebox(10,1){0.8512}}\\\colorbox{white}{\makebox(10,1){0.8510}}\\\colorbox{white}{\makebox(10,1){0.8514}}\\\colorbox{white}{\makebox(10,1){0.8509}}\\\colorbox{white}{\makebox(10,1){0.8497}}\\\colorbox{white}{\makebox(10,1){0.8511}}\\\colorbox{green!30}{\makebox(10,1){\textbf{\textcolor{blue}{0.8523}}}}\\\colorbox{white}{\makebox(10,1){0.8511}}\\\colorbox{white}{\makebox(10,1){0.8508}}\\\colorbox{white}{\makebox(10,1){0.8516}}\\\colorbox{green!30}{\makebox(10,1){0.8516}}\\\colorbox{green!30}{\makebox(10,1){0.8522}}\\\colorbox{white}{\makebox(10,1){0.8515}}\\\colorbox{white}{\makebox(10,1){0.8512}}\\\colorbox{white}{\makebox(10,1){0.8511}}\\\colorbox{white}{\makebox(10,1){0.8510}}\\\colorbox{white}{\makebox(10,1){0.8511}}\\\colorbox{white}{\makebox(10,1){0.8501}}\\\colorbox{white}{\makebox(10,1){0.8512}}\\\colorbox{white}{\makebox(10,1){0.8518}}}&
\makecell{\colorbox{white}{\makebox(10,1){0.8174}}\\\colorbox{white}{\makebox(10,1){0.8174}}\\\colorbox{white}{\makebox(10,1){0.8176}}\\\colorbox{white}{\makebox(10,1){0.8171}}\\\colorbox{white}{\makebox(10,1){0.8149}}\\\colorbox{white}{\makebox(10,1){0.8173}}\\\textbf{\textcolor{red}{0.8188}}\\\colorbox{white}{\makebox(10,1){0.8174}}\\\colorbox{white}{\makebox(10,1){0.8170}}\\\colorbox{white}{\makebox(10,1){0.8179}}\\\colorbox{white}{\makebox(10,1){0.8181}}\\\colorbox{white}{\makebox(10,1){0.8186}}\\\colorbox{white}{\makebox(10,1){0.8178}}\\\colorbox{white}{\makebox(10,1){0.8174}}\\\colorbox{white}{\makebox(10,1){0.8173}}\\\colorbox{white}{\makebox(10,1){0.8171}}\\\colorbox{white}{\makebox(10,1){0.8173}}\\\colorbox{white}{\makebox(10,1){0.8158}}\\\colorbox{white}{\makebox(10,1){0.8177}}\\\colorbox{white}{\makebox(10,1){0.8181}}}
\\ \midrule
\makecell{0.2} &
\makecell{\colorbox{white}{\makebox(10,1){CLAHE}}\\
\colorbox{white}{\makebox(10,1){Coarse Dropout}}\\
\colorbox{white}{\makebox(10,1){Elastic Transform}}\\
\colorbox{white}{\makebox(10,1){Emboss}}\\
\colorbox{white}{\makebox(10,1){Flip}}\\
\colorbox{white}{\makebox(10,1){Gaussian Blur}}\\
\colorbox{white}{\makebox(10,1){Grid Distortion}}\\
\colorbox{white}{\makebox(10,1){Grid Dropout}}\\
\colorbox{white}{\makebox(10,1){Image Compression}}\\
\colorbox{white}{\makebox(10,1){Median Blur}}\\
\colorbox{white}{\makebox(10,1){Optical Distortion}}\\
\colorbox{white}{\makebox(10,1){Piecewise Affine}}\\
\colorbox{white}{\makebox(10,1){Posterize}}\\
\colorbox{white}{\makebox(10,1){\gls{rbc}}}\\
\colorbox{white}{\makebox(10,1){Random Crop}}\\
\colorbox{white}{\makebox(10,1){Random Gamma}}\\
\colorbox{white}{\makebox(10,1){Random Snow}}\\
\colorbox{white}{\makebox(10,1){Rotate}}\\
\colorbox{white}{\makebox(10,1){Sharpen}}\\
\colorbox{white}{\makebox(10,1){Shift Scale Rotate}}}&
\makecell{\colorbox{white}{\makebox(10,1){0.6408}}\\\colorbox{white}{\makebox(10,1){0.6420}}\\\colorbox{green!30}{\makebox(10,1){0.6526}}\\\colorbox{white}{\makebox(10,1){0.6430}}\\\colorbox{green!30}{\makebox(10,1){0.6501}}\\\colorbox{white}{\makebox(10,1){0.6431}}\\\colorbox{green!30}{\makebox(10,1){\textbf{\textcolor{blue}{0.6543}}}}\\\colorbox{white}{\makebox(10,1){0.6408}}\\\colorbox{white}{\makebox(10,1){0.6460}}\\\colorbox{white}{\makebox(10,1){0.6467}}\\\colorbox{green!30}{\makebox(10,1){0.6492}}\\\colorbox{green!30}{\makebox(10,1){0.6482}}\\\colorbox{white}{\makebox(10,1){0.6472}}\\\colorbox{white}{\makebox(10,1){0.6415}}\\\colorbox{white}{\makebox(10,1){0.6449}}\\\colorbox{white}{\makebox(10,1){0.6459}}\\\colorbox{white}{\makebox(10,1){0.6416}}\\\colorbox{green!30}{\makebox(10,1){0.6526}}\\\colorbox{white}{\makebox(10,1){0.6466}}\\\colorbox{green!30}{\makebox(10,1){0.6534}}}&
\makecell{\colorbox{white}{\makebox(10,1){0.6005}}\\\colorbox{white}{\makebox(10,1){0.6014}}\\\colorbox{white}{\makebox(10,1){0.6121}}\\\colorbox{white}{\makebox(10,1){0.6026}}\\\colorbox{white}{\makebox(10,1){0.6091}}\\\colorbox{white}{\makebox(10,1){0.6026}}\\\textbf{\textcolor{red}{0.6139}}\\\colorbox{white}{\makebox(10,1){0.6006}}\\\colorbox{white}{\makebox(10,1){0.6057}}\\\colorbox{white}{\makebox(10,1){0.6063}}\\\colorbox{white}{\makebox(10,1){0.6085}}\\\colorbox{white}{\makebox(10,1){0.6079}}\\\colorbox{white}{\makebox(10,1){0.6063}}\\\colorbox{white}{\makebox(10,1){0.6008}}\\\colorbox{white}{\makebox(10,1){0.6049}}\\\colorbox{white}{\makebox(10,1){0.6051}}\\\colorbox{white}{\makebox(10,1){0.6018}}\\\colorbox{white}{\makebox(10,1){0.6121}}\\\colorbox{white}{\makebox(10,1){0.6063}}\\\colorbox{white}{\makebox(10,1){0.6127}}}&
\makecell{\colorbox{white}{\makebox(10,1){0.4764}}\\\colorbox{white}{\makebox(10,1){0.4776}}\\\colorbox{green!30}{\makebox(10,1){0.4840}}\\\colorbox{white}{\makebox(10,1){0.4790}}\\\colorbox{green!30}{\makebox(10,1){0.4813}}\\\colorbox{white}{\makebox(10,1){0.4775}}\\\colorbox{green!30}{\makebox(10,1){0.4835}}\\\colorbox{white}{\makebox(10,1){0.4778}}\\\colorbox{white}{\makebox(10,1){0.4782}}\\\colorbox{white}{\makebox(10,1){0.4769}}\\\colorbox{green!30}{\makebox(10,1){0.4821}}\\\colorbox{green!30}{\makebox(10,1){0.4815}}\\\colorbox{white}{\makebox(10,1){0.4793}}\\\colorbox{white}{\makebox(10,1){0.4791}}\\\colorbox{white}{\makebox(10,1){0.4771}}\\\colorbox{white}{\makebox(10,1){0.4772}}\\\colorbox{white}{\makebox(10,1){0.4752}}\\\colorbox{green!30}{\makebox(10,1){0.4831}}\\\colorbox{white}{\makebox(10,1){0.4778}}\\\colorbox{green!30}{\makebox(10,1){\textbf{\textcolor{blue}{0.4843}}}}}&
\makecell{\colorbox{white}{\makebox(10,1){0.4293}}\\\colorbox{white}{\makebox(10,1){0.4307}}\\\colorbox{white}{\makebox(10,1){0.4366}}\\\colorbox{white}{\makebox(10,1){0.4316}}\\\colorbox{white}{\makebox(10,1){0.4341}}\\\colorbox{white}{\makebox(10,1){0.4300}}\\\colorbox{white}{\makebox(10,1){0.4365}}\\\colorbox{white}{\makebox(10,1){0.4308}}\\\colorbox{white}{\makebox(10,1){0.4308}}\\\colorbox{white}{\makebox(10,1){0.4302}}\\\colorbox{white}{\makebox(10,1){0.4346}}\\\colorbox{white}{\makebox(10,1){0.4343}}\\\colorbox{white}{\makebox(10,1){0.4318}}\\\colorbox{white}{\makebox(10,1){0.4312}}\\\colorbox{white}{\makebox(10,1){0.4299}}\\\colorbox{white}{\makebox(10,1){0.4301}}\\\colorbox{white}{\makebox(10,1){0.4280}}\\\colorbox{white}{\makebox(10,1){0.4357}}\\\colorbox{white}{\makebox(10,1){0.4310}}\\\textbf{\textcolor{red}{0.4369}}}&
\makecell{\colorbox{white}{\makebox(10,1){0.7919}}\\\colorbox{white}{\makebox(10,1){0.7952}}\\\colorbox{green!30}{\makebox(10,1){\textbf{\textcolor{blue}{0.8137}}}}\\\colorbox{white}{\makebox(10,1){0.7990}}\\\colorbox{green!30}{\makebox(10,1){0.8066}}\\\colorbox{white}{\makebox(10,1){0.7920}}\\\colorbox{green!30}{\makebox(10,1){0.8102}}\\\colorbox{white}{\makebox(10,1){0.7950}}\\\colorbox{white}{\makebox(10,1){0.7962}}\\\colorbox{white}{\makebox(10,1){0.7964}}\\\colorbox{green!30}{\makebox(10,1){0.8057}}\\\colorbox{green!30}{\makebox(10,1){0.8106}}\\\colorbox{white}{\makebox(10,1){0.7930}}\\\colorbox{white}{\makebox(10,1){0.7995}}\\\colorbox{white}{\makebox(10,1){0.7963}}\\\colorbox{white}{\makebox(10,1){0.7929}}\\\colorbox{white}{\makebox(10,1){0.7961}}\\\colorbox{green!30}{\makebox(10,1){0.8119}}\\\colorbox{white}{\makebox(10,1){0.7867}}\\\colorbox{green!30}{\makebox(10,1){0.8135}}}&
\makecell{\colorbox{white}{\makebox(10,1){0.7273}}\\\colorbox{white}{\makebox(10,1){0.7303}}\\\textbf{\textcolor{red}{0.7469}}\\\colorbox{white}{\makebox(10,1){0.7326}}\\\colorbox{white}{\makebox(10,1){0.7408}}\\\colorbox{white}{\makebox(10,1){0.7276}}\\\colorbox{white}{\makebox(10,1){0.7441}}\\\colorbox{white}{\makebox(10,1){0.7298}}\\\colorbox{white}{\makebox(10,1){0.7310}}\\\colorbox{white}{\makebox(10,1){0.7305}}\\\colorbox{white}{\makebox(10,1){0.7396}}\\\colorbox{white}{\makebox(10,1){0.7444}}\\\colorbox{white}{\makebox(10,1){0.7280}}\\\colorbox{white}{\makebox(10,1){0.7333}}\\\colorbox{white}{\makebox(10,1){0.7309}}\\\colorbox{white}{\makebox(10,1){0.7274}}\\\colorbox{white}{\makebox(10,1){0.7291}}\\\colorbox{white}{\makebox(10,1){0.7456}}\\\colorbox{white}{\makebox(10,1){0.7209}}\\\colorbox{white}{\makebox(10,1){0.7462}}}&
\makecell{\colorbox{white}{\makebox(10,1){0.9025}}\\\colorbox{white}{\makebox(10,1){0.9033}}\\\colorbox{white}{\makebox(10,1){0.9012}}\\\colorbox{green!30}{\makebox(10,1){0.9043}}\\\colorbox{white}{\makebox(10,1){0.8977}}\\\colorbox{white}{\makebox(10,1){0.9021}}\\\colorbox{white}{\makebox(10,1){0.9007}}\\\colorbox{white}{\makebox(10,1){0.9019}}\\\colorbox{green!30}{\makebox(10,1){\textbf{\textcolor{blue}{0.9048}}}}\\\colorbox{white}{\makebox(10,1){0.9022}}\\\colorbox{white}{\makebox(10,1){0.9030}}\\\colorbox{white}{\makebox(10,1){0.9024}}\\\colorbox{white}{\makebox(10,1){0.9027}}\\\colorbox{white}{\makebox(10,1){0.9019}}\\\colorbox{white}{\makebox(10,1){0.9034}}\\\colorbox{white}{\makebox(10,1){0.9025}}\\\colorbox{white}{\makebox(10,1){0.9028}}\\\colorbox{white}{\makebox(10,1){0.8945}}\\\colorbox{white}{\makebox(10,1){0.9035}}\\\colorbox{white}{\makebox(10,1){0.8964}}}&
\makecell{\colorbox{white}{\makebox(10,1){0.8461}}\\\colorbox{white}{\makebox(10,1){0.8471}}\\\colorbox{white}{\makebox(10,1){0.8439}}\\\colorbox{white}{\makebox(10,1){0.8484}}\\\colorbox{white}{\makebox(10,1){0.8397}}\\\colorbox{white}{\makebox(10,1){0.8457}}\\\colorbox{white}{\makebox(10,1){0.8432}}\\\colorbox{white}{\makebox(10,1){0.8452}}\\\textbf{\textcolor{red}{0.8494}}\\\colorbox{white}{\makebox(10,1){0.8455}}\\\colorbox{white}{\makebox(10,1){0.8466}}\\\colorbox{white}{\makebox(10,1){0.8454}}\\\colorbox{white}{\makebox(10,1){0.8463}}\\\colorbox{white}{\makebox(10,1){0.8453}}\\\colorbox{white}{\makebox(10,1){0.8474}}\\\colorbox{white}{\makebox(10,1){0.8462}}\\\colorbox{white}{\makebox(10,1){0.8467}}\\\colorbox{white}{\makebox(10,1){0.8362}}\\\colorbox{white}{\makebox(10,1){0.8474}}\\\colorbox{white}{\makebox(10,1){0.8385}}}&
\makecell{\colorbox{white}{\makebox(10,1){0.8512}}\\\colorbox{white}{\makebox(10,1){0.8512}}\\\colorbox{white}{\makebox(10,1){0.8517}}\\\colorbox{white}{\makebox(10,1){0.8510}}\\\colorbox{white}{\makebox(10,1){0.8486}}\\\colorbox{white}{\makebox(10,1){0.8509}}\\\colorbox{green!30}{\makebox(10,1){\textbf{\textcolor{blue}{0.8524}}}}\\\colorbox{white}{\makebox(10,1){0.8507}}\\\colorbox{white}{\makebox(10,1){0.8512}}\\\colorbox{white}{\makebox(10,1){0.8510}}\\\colorbox{green!30}{\makebox(10,1){0.8518}}\\\colorbox{white}{\makebox(10,1){0.8516}}\\\colorbox{white}{\makebox(10,1){0.8512}}\\\colorbox{white}{\makebox(10,1){0.8512}}\\\colorbox{white}{\makebox(10,1){0.8513}}\\\colorbox{white}{\makebox(10,1){0.8512}}\\\colorbox{white}{\makebox(10,1){0.8507}}\\\colorbox{white}{\makebox(10,1){0.8498}}\\\colorbox{white}{\makebox(10,1){0.8515}}\\\colorbox{white}{\makebox(10,1){0.8517}}}&
\makecell{\colorbox{white}{\makebox(10,1){0.8173}}\\\colorbox{white}{\makebox(10,1){0.8176}}\\\colorbox{white}{\makebox(10,1){0.8178}}\\\colorbox{white}{\makebox(10,1){0.8170}}\\\colorbox{white}{\makebox(10,1){0.8135}}\\\colorbox{white}{\makebox(10,1){0.8171}}\\\textbf{\textcolor{red}{0.8189}}\\\colorbox{white}{\makebox(10,1){0.8167}}\\\colorbox{white}{\makebox(10,1){0.8175}}\\\colorbox{white}{\makebox(10,1){0.8173}}\\\colorbox{white}{\makebox(10,1){0.8183}}\\\colorbox{white}{\makebox(10,1){0.8179}}\\\colorbox{white}{\makebox(10,1){0.8174}}\\\colorbox{white}{\makebox(10,1){0.8174}}\\\colorbox{white}{\makebox(10,1){0.8175}}\\\colorbox{white}{\makebox(10,1){0.8173}}\\\colorbox{white}{\makebox(10,1){0.8166}}\\\colorbox{white}{\makebox(10,1){0.8153}}\\\colorbox{white}{\makebox(10,1){0.8180}}\\\colorbox{white}{\makebox(10,1){0.8180}}}
\\ \bottomrule
\end{tabular}
}}
\label{tab:tests4}
\end{table}

In the second evaluation of data augmentations, the architecture was trained for 100 epochs with a learning rate of 0.001. The learning rate was divided by 10 every 20 epochs. Two probabilities of applying the data augmentation were evaluated: 0.1 and 0.2. As presented in Table~\ref{tab:tests4}, the MosMed achieved the most significant improvements, with the Grid Distortion with probability 0.1 and the Elastic Transform with probability 0.2 increasing the F-score by 2\% compared with the baseline. However, unlike the first evaluation, the MosMeg achieved better F-scores in only seven augmentations instead of fourteen augmentations.

\begin{table}[!ht]
\centering
\caption[Results of the third data augmentation evaluation.]{Results of the third data augmentation evaluation in the test set. The blue-colored values indicate the best F-scores, and the red-colored values indicate the best \glspl*{iou} values. Two probabilities of applying the data augmentation were evaluated: 0.1 and 0.2. The green highlighted values show the data augmentations in which the P-value achieved values lower than 0.05, and the null hypothesis was rejected.} 
\resizebox{\textwidth}{!}{%
\tiny{
\begin{tabular}{cccccccccccc}

\toprule
\multicolumn{1}{c}{\textbf{Probability}} &
\multicolumn{1}{c}{\textbf{Augmentation}} &
\multicolumn{2}{c}{\textbf{CC-CCII}} &
\multicolumn{2}{c}{\textbf{MedSeg}} &
\multicolumn{2}{c}{\textbf{MosMed}} &
\multicolumn{2}{c}{\textbf{Ricord1a}} &
\multicolumn{2}{c}{\textbf{Zenodo}} 
\\ \midrule

\makecell{} &
\makecell{} &
\makecell{F-score} &
\makecell{\gls*{iou}} & 
\makecell{F-score} &
\makecell{\gls*{iou}} & 
\makecell{F-score} &
\makecell{\gls*{iou}} & 
\makecell{F-score} &
\makecell{\gls*{iou}} & 
\makecell{F-score} &
\makecell{\gls*{iou}}
\\ \midrule

\makecell{} &
\makecell{No Augmentation}&
\makecell{0.6317}&
\makecell{0.5893}&
\makecell{0.4658}&
\makecell{0.4183}&
\makecell{0.7736}&
\makecell{0.7089}&
\makecell{0.9133}&
\makecell{0.8610}&
\makecell{0.8496}&
\makecell{0.8153}

\\ \midrule

\makecell{0.1} &
\makecell{\colorbox{white}{\makebox(10,1){CLAHE}}\\
\colorbox{white}{\makebox(10,1){Coarse Dropout}}\\
\colorbox{white}{\makebox(10,1){Elastic Transform}}\\
\colorbox{white}{\makebox(10,1){Emboss}}\\
\colorbox{white}{\makebox(10,1){Flip}}\\
\colorbox{white}{\makebox(10,1){Gaussian Blur}}\\
\colorbox{white}{\makebox(10,1){Grid Distortion}}\\
\colorbox{white}{\makebox(10,1){Grid Dropout}}\\
\colorbox{white}{\makebox(10,1){Image Compression}}\\
\colorbox{white}{\makebox(10,1){Median Blur}}\\
\colorbox{white}{\makebox(10,1){Optical Distortion}}\\
\colorbox{white}{\makebox(10,1){Piecewise Affine}}\\
\colorbox{white}{\makebox(10,1){Posterize}}\\
\colorbox{white}{\makebox(10,1){\gls{rbc}}}\\
\colorbox{white}{\makebox(10,1){Random Crop}}\\
\colorbox{white}{\makebox(10,1){Random Gamma}}\\
\colorbox{white}{\makebox(10,1){Random Snow}}\\
\colorbox{white}{\makebox(10,1){Rotate}}\\
\colorbox{white}{\makebox(10,1){Sharpen}}\\
\colorbox{white}{\makebox(10,1){Shift Scale Rotate}}}&
\makecell{\colorbox{white}{\makebox(10,1){0.6288}}\\\colorbox{white}{\makebox(10,1){0.6275}}\\\colorbox{green!30}{\makebox(10,1){0.6330}}\\\colorbox{white}{\makebox(10,1){0.6282}}\\\colorbox{white}{\makebox(10,1){0.6348}}\\\colorbox{white}{\makebox(10,1){0.6310}}\\\colorbox{white}{\makebox(10,1){0.6328}}\\\colorbox{white}{\makebox(10,1){0.6321}}\\\colorbox{white}{\makebox(10,1){0.6296}}\\\colorbox{white}{\makebox(10,1){0.6311}}\\\colorbox{white}{\makebox(10,1){0.6328}}\\\colorbox{green!30}{\makebox(10,1){0.6339}}\\\colorbox{white}{\makebox(10,1){0.6308}}\\\colorbox{white}{\makebox(10,1){0.6275}}\\\colorbox{white}{\makebox(10,1){0.6295}}\\\colorbox{white}{\makebox(10,1){0.6301}}\\\colorbox{white}{\makebox(10,1){0.6278}}\\\colorbox{green!30}{\makebox(10,1){0.6350}}\\\colorbox{white}{\makebox(10,1){0.6322}}\\\colorbox{green!30}{\makebox(10,1){\textbf{\textcolor{blue}{0.6381}}}}}&
\makecell{\colorbox{white}{\makebox(10,1){0.5873}}\\\colorbox{white}{\makebox(10,1){0.5863}}\\\colorbox{white}{\makebox(10,1){0.5912}}\\\colorbox{white}{\makebox(10,1){0.5863}}\\\colorbox{white}{\makebox(10,1){0.5922}}\\\colorbox{white}{\makebox(10,1){0.5890}}\\\colorbox{white}{\makebox(10,1){0.5911}}\\\colorbox{white}{\makebox(10,1){0.5900}}\\\colorbox{white}{\makebox(10,1){0.5876}}\\\colorbox{white}{\makebox(10,1){0.5886}}\\\colorbox{white}{\makebox(10,1){0.5911}}\\\colorbox{white}{\makebox(10,1){0.5926}}\\\colorbox{white}{\makebox(10,1){0.5889}}\\\colorbox{white}{\makebox(10,1){0.5858}}\\\colorbox{white}{\makebox(10,1){0.5874}}\\\colorbox{white}{\makebox(10,1){0.5883}}\\\colorbox{white}{\makebox(10,1){0.5863}}\\\colorbox{white}{\makebox(10,1){0.5929}}\\\colorbox{white}{\makebox(10,1){0.5895}}\\\textbf{\textcolor{red}{0.5961}}}&
\makecell{\colorbox{white}{\makebox(10,1){0.4618}}\\\colorbox{white}{\makebox(10,1){0.4647}}\\\colorbox{white}{\makebox(10,1){0.4679}}\\\colorbox{white}{\makebox(10,1){0.4654}}\\\colorbox{green!30}{\makebox(10,1){0.4686}}\\\colorbox{white}{\makebox(10,1){0.4675}}\\\colorbox{white}{\makebox(10,1){0.4704}}\\\colorbox{white}{\makebox(10,1){0.4637}}\\\colorbox{white}{\makebox(10,1){0.4653}}\\\colorbox{white}{\makebox(10,1){0.4659}}\\\colorbox{white}{\makebox(10,1){0.4696}}\\\colorbox{white}{\makebox(10,1){0.4675}}\\\colorbox{white}{\makebox(10,1){0.4645}}\\\colorbox{white}{\makebox(10,1){0.4628}}\\\colorbox{white}{\makebox(10,1){0.4626}}\\\colorbox{white}{\makebox(10,1){0.4627}}\\\colorbox{white}{\makebox(10,1){0.4610}}\\\colorbox{green!30}{\makebox(10,1){\textbf{\textcolor{blue}{0.4719}}}}\\\colorbox{white}{\makebox(10,1){0.4650}}\\\colorbox{green!30}{\makebox(10,1){0.4710}}}&
\makecell{\colorbox{white}{\makebox(10,1){0.4149}}\\\colorbox{white}{\makebox(10,1){0.4179}}\\\colorbox{white}{\makebox(10,1){0.4208}}\\\colorbox{white}{\makebox(10,1){0.4182}}\\\colorbox{white}{\makebox(10,1){0.4214}}\\\colorbox{white}{\makebox(10,1){0.4202}}\\\colorbox{white}{\makebox(10,1){0.4229}}\\\colorbox{white}{\makebox(10,1){0.4171}}\\\colorbox{white}{\makebox(10,1){0.4182}}\\\colorbox{white}{\makebox(10,1){0.4193}}\\\colorbox{white}{\makebox(10,1){0.4219}}\\\colorbox{white}{\makebox(10,1){0.4206}}\\\colorbox{white}{\makebox(10,1){0.4179}}\\\colorbox{white}{\makebox(10,1){0.4159}}\\\colorbox{white}{\makebox(10,1){0.4156}}\\\colorbox{white}{\makebox(10,1){0.4159}}\\\colorbox{white}{\makebox(10,1){0.4146}}\\\textbf{\textcolor{red}{0.4243}}\\\colorbox{white}{\makebox(10,1){0.4177}}\\\colorbox{white}{\makebox(10,1){0.4236}}}&
\makecell{\colorbox{white}{\makebox(10,1){0.7698}}\\\colorbox{green!30}{\makebox(10,1){0.7831}}\\\colorbox{green!30}{\makebox(10,1){0.7907}}\\\colorbox{white}{\makebox(10,1){0.7758}}\\\colorbox{green!30}{\makebox(10,1){0.7933}}\\\colorbox{white}{\makebox(10,1){0.7751}}\\\colorbox{green!30}{\makebox(10,1){0.7907}}\\\colorbox{green!30}{\makebox(10,1){0.7809}}\\\colorbox{white}{\makebox(10,1){0.7725}}\\\colorbox{white}{\makebox(10,1){0.7720}}\\\colorbox{green!30}{\makebox(10,1){0.7862}}\\\colorbox{green!30}{\makebox(10,1){0.7923}}\\\colorbox{green!30}{\makebox(10,1){0.7762}}\\\colorbox{white}{\makebox(10,1){0.7674}}\\\colorbox{green!30}{\makebox(10,1){0.7764}}\\\colorbox{white}{\makebox(10,1){0.7744}}\\\colorbox{white}{\makebox(10,1){0.7675}}\\\colorbox{green!30}{\makebox(10,1){0.7922}}\\\colorbox{green!30}{\makebox(10,1){0.7740}}\\\colorbox{green!30}{\makebox(10,1){\textbf{\textcolor{blue}{0.7962}}}}}&
\makecell{\colorbox{white}{\makebox(10,1){0.7068}}\\\colorbox{white}{\makebox(10,1){0.7178}}\\\colorbox{white}{\makebox(10,1){0.7249}}\\\colorbox{white}{\makebox(10,1){0.7108}}\\\colorbox{white}{\makebox(10,1){0.7278}}\\\colorbox{white}{\makebox(10,1){0.7110}}\\\colorbox{white}{\makebox(10,1){0.7241}}\\\colorbox{white}{\makebox(10,1){0.7158}}\\\colorbox{white}{\makebox(10,1){0.7086}}\\\colorbox{white}{\makebox(10,1){0.7088}}\\\colorbox{white}{\makebox(10,1){0.7205}}\\\colorbox{white}{\makebox(10,1){0.7266}}\\\colorbox{white}{\makebox(10,1){0.7113}}\\\colorbox{white}{\makebox(10,1){0.7029}}\\\colorbox{white}{\makebox(10,1){0.7112}}\\\colorbox{white}{\makebox(10,1){0.7099}}\\\colorbox{white}{\makebox(10,1){0.7032}}\\\colorbox{white}{\makebox(10,1){0.7268}}\\\colorbox{white}{\makebox(10,1){0.7099}}\\\textbf{\textcolor{red}{0.7298}}}&
\makecell{\colorbox{white}{\makebox(10,1){0.9140}}\\\colorbox{green!30}{\makebox(10,1){0.9142}}\\\colorbox{white}{\makebox(10,1){0.9136}}\\\colorbox{green!30}{\makebox(10,1){0.9139}}\\\colorbox{white}{\makebox(10,1){0.9130}}\\\colorbox{green!30}{\makebox(10,1){0.9145}}\\\colorbox{white}{\makebox(10,1){0.9144}}\\\colorbox{white}{\makebox(10,1){0.9139}}\\\colorbox{green!30}{\makebox(10,1){0.9142}}\\\colorbox{green!30}{\makebox(10,1){0.9141}}\\\colorbox{green!30}{\makebox(10,1){0.9141}}\\\colorbox{green!30}{\makebox(10,1){\textbf{\textcolor{blue}{0.9147}}}}\\\colorbox{green!30}{\makebox(10,1){0.9141}}\\\colorbox{white}{\makebox(10,1){0.9131}}\\\colorbox{white}{\makebox(10,1){0.9128}}\\\colorbox{white}{\makebox(10,1){0.9139}}\\\colorbox{green!30}{\makebox(10,1){0.9141}}\\\colorbox{white}{\makebox(10,1){0.9135}}\\\colorbox{green!30}{\makebox(10,1){0.9146}}\\\colorbox{white}{\makebox(10,1){0.9139}}}&
\makecell{\colorbox{white}{\makebox(10,1){0.8617}}\\\colorbox{white}{\makebox(10,1){0.8620}}\\\colorbox{white}{\makebox(10,1){0.8610}}\\\colorbox{white}{\makebox(10,1){0.8618}}\\\colorbox{white}{\makebox(10,1){0.8603}}\\\colorbox{white}{\makebox(10,1){0.8625}}\\\colorbox{white}{\makebox(10,1){0.8621}}\\\colorbox{white}{\makebox(10,1){0.8616}}\\\colorbox{white}{\makebox(10,1){0.8620}}\\\colorbox{white}{\makebox(10,1){0.8620}}\\\colorbox{white}{\makebox(10,1){0.8620}}\\\colorbox{white}{\makebox(10,1){0.8625}}\\\colorbox{white}{\makebox(10,1){0.8621}}\\\colorbox{white}{\makebox(10,1){0.8607}}\\\colorbox{white}{\makebox(10,1){0.8605}}\\\colorbox{white}{\makebox(10,1){0.8616}}\\\colorbox{white}{\makebox(10,1){0.8620}}\\\colorbox{white}{\makebox(10,1){0.8609}}\\\textbf{\textcolor{red}{0.8626}}\\\colorbox{white}{\makebox(10,1){0.8614}}}&
\makecell{\colorbox{white}{\makebox(10,1){0.8498}}\\\colorbox{white}{\makebox(10,1){0.8496}}\\\colorbox{green!30}{\makebox(10,1){0.8505}}\\\colorbox{white}{\makebox(10,1){0.8499}}\\\colorbox{white}{\makebox(10,1){0.8493}}\\\colorbox{green!30}{\makebox(10,1){0.8498}}\\\colorbox{green!30}{\makebox(10,1){\textbf{\textcolor{blue}{0.8509}}}}\\\colorbox{white}{\makebox(10,1){0.8495}}\\\colorbox{white}{\makebox(10,1){0.8477}}\\\colorbox{green!30}{\makebox(10,1){0.8504}}\\\colorbox{white}{\makebox(10,1){0.8496}}\\\colorbox{green!30}{\makebox(10,1){0.8504}}\\\colorbox{green!30}{\makebox(10,1){0.8502}}\\\colorbox{white}{\makebox(10,1){0.8497}}\\\colorbox{white}{\makebox(10,1){0.8498}}\\\colorbox{white}{\makebox(10,1){0.8494}}\\\colorbox{white}{\makebox(10,1){0.8496}}\\\colorbox{white}{\makebox(10,1){0.8490}}\\\colorbox{white}{\makebox(10,1){0.8499}}\\\colorbox{green!30}{\makebox(10,1){0.8506}}}&
\makecell{\colorbox{white}{\makebox(10,1){0.8155}}\\\colorbox{white}{\makebox(10,1){0.8152}}\\\colorbox{white}{\makebox(10,1){0.8163}}\\\colorbox{white}{\makebox(10,1){0.8157}}\\\colorbox{white}{\makebox(10,1){0.8147}}\\\colorbox{white}{\makebox(10,1){0.8156}}\\\textbf{\textcolor{red}{0.8169}}\\\colorbox{white}{\makebox(10,1){0.8152}}\\\colorbox{white}{\makebox(10,1){0.8144}}\\\colorbox{white}{\makebox(10,1){0.8165}}\\\colorbox{white}{\makebox(10,1){0.8160}}\\\colorbox{white}{\makebox(10,1){0.8163}}\\\colorbox{white}{\makebox(10,1){0.8159}}\\\colorbox{white}{\makebox(10,1){0.8154}}\\\colorbox{white}{\makebox(10,1){0.8156}}\\\colorbox{white}{\makebox(10,1){0.8150}}\\\colorbox{white}{\makebox(10,1){0.8151}}\\\colorbox{white}{\makebox(10,1){0.8143}}\\\colorbox{white}{\makebox(10,1){0.8157}}\\\colorbox{white}{\makebox(10,1){0.8165}}}
\\ \midrule
\makecell{0.2} &
\makecell{\colorbox{white}{\makebox(10,1){CLAHE}}\\
\colorbox{white}{\makebox(10,1){Coarse Dropout}}\\
\colorbox{white}{\makebox(10,1){Elastic Transform}}\\
\colorbox{white}{\makebox(10,1){Emboss}}\\
\colorbox{white}{\makebox(10,1){Flip}}\\
\colorbox{white}{\makebox(10,1){Gaussian Blur}}\\
\colorbox{white}{\makebox(10,1){Grid Distortion}}\\
\colorbox{white}{\makebox(10,1){Grid Dropout}}\\
\colorbox{white}{\makebox(10,1){Image Compression}}\\
\colorbox{white}{\makebox(10,1){Median Blur}}\\
\colorbox{white}{\makebox(10,1){Optical Distortion}}\\
\colorbox{white}{\makebox(10,1){Piecewise Affine}}\\
\colorbox{white}{\makebox(10,1){Posterize}}\\
\colorbox{white}{\makebox(10,1){\gls{rbc}}}\\
\colorbox{white}{\makebox(10,1){Random Crop}}\\
\colorbox{white}{\makebox(10,1){Random Gamma}}\\
\colorbox{white}{\makebox(10,1){Random Snow}}\\
\colorbox{white}{\makebox(10,1){Rotate}}\\
\colorbox{white}{\makebox(10,1){Sharpen}}\\
\colorbox{white}{\makebox(10,1){Shift Scale Rotate}}}&
\makecell{\colorbox{white}{\makebox(10,1){0.6272}}\\\colorbox{white}{\makebox(10,1){0.6286}}\\\colorbox{green!30}{\makebox(10,1){0.6373}}\\\colorbox{white}{\makebox(10,1){0.6288}}\\\colorbox{green!30}{\makebox(10,1){0.6365}}\\\colorbox{white}{\makebox(10,1){0.6297}}\\\colorbox{white}{\makebox(10,1){0.6370}}\\\colorbox{white}{\makebox(10,1){0.6304}}\\\colorbox{white}{\makebox(10,1){0.6312}}\\\colorbox{white}{\makebox(10,1){0.6297}}\\\colorbox{green!30}{\makebox(10,1){0.6335}}\\\colorbox{green!30}{\makebox(10,1){0.6344}}\\\colorbox{white}{\makebox(10,1){0.6303}}\\\colorbox{white}{\makebox(10,1){0.6273}}\\\colorbox{white}{\makebox(10,1){0.6315}}\\\colorbox{white}{\makebox(10,1){0.6313}}\\\colorbox{white}{\makebox(10,1){0.6258}}\\\colorbox{green!30}{\makebox(10,1){0.6368}}\\\colorbox{white}{\makebox(10,1){0.6323}}\\\colorbox{green!30}{\makebox(10,1){\textbf{\textcolor{blue}{0.6380}}}}}&
\makecell{\colorbox{white}{\makebox(10,1){0.5854}}\\\colorbox{white}{\makebox(10,1){0.5870}}\\\colorbox{white}{\makebox(10,1){0.5956}}\\\colorbox{white}{\makebox(10,1){0.5870}}\\\colorbox{white}{\makebox(10,1){0.5944}}\\\colorbox{white}{\makebox(10,1){0.5880}}\\\colorbox{white}{\makebox(10,1){0.5951}}\\\colorbox{white}{\makebox(10,1){0.5887}}\\\colorbox{white}{\makebox(10,1){0.5888}}\\\colorbox{white}{\makebox(10,1){0.5882}}\\\colorbox{white}{\makebox(10,1){0.5914}}\\\colorbox{white}{\makebox(10,1){0.5927}}\\\colorbox{white}{\makebox(10,1){0.5886}}\\\colorbox{white}{\makebox(10,1){0.5860}}\\\colorbox{white}{\makebox(10,1){0.5894}}\\\colorbox{white}{\makebox(10,1){0.5893}}\\\colorbox{white}{\makebox(10,1){0.5841}}\\\colorbox{white}{\makebox(10,1){0.5948}}\\\colorbox{white}{\makebox(10,1){0.5901}}\\\textbf{\textcolor{red}{0.5960}}}&
\makecell{\colorbox{white}{\makebox(10,1){0.4649}}\\\colorbox{white}{\makebox(10,1){0.4625}}\\\colorbox{green!30}{\makebox(10,1){0.4714}}\\\colorbox{white}{\makebox(10,1){0.4639}}\\\colorbox{green!30}{\makebox(10,1){0.4696}}\\\colorbox{white}{\makebox(10,1){0.4659}}\\\colorbox{white}{\makebox(10,1){0.4725}}\\\colorbox{white}{\makebox(10,1){0.4649}}\\\colorbox{white}{\makebox(10,1){0.4661}}\\\colorbox{white}{\makebox(10,1){0.4681}}\\\colorbox{green!30}{\makebox(10,1){0.4691}}\\\colorbox{green!30}{\makebox(10,1){0.4726}}\\\colorbox{white}{\makebox(10,1){0.4647}}\\\colorbox{white}{\makebox(10,1){0.4668}}\\\colorbox{white}{\makebox(10,1){0.4625}}\\\colorbox{white}{\makebox(10,1){0.4629}}\\\colorbox{white}{\makebox(10,1){0.4599}}\\\colorbox{green!30}{\makebox(10,1){\textbf{\textcolor{blue}{0.4739}}}}\\\colorbox{white}{\makebox(10,1){0.4668}}\\\colorbox{green!30}{\makebox(10,1){0.4712}}}&
\makecell{\colorbox{white}{\makebox(10,1){0.4177}}\\\colorbox{white}{\makebox(10,1){0.4160}}\\\colorbox{white}{\makebox(10,1){0.4235}}\\\colorbox{white}{\makebox(10,1){0.4171}}\\\colorbox{white}{\makebox(10,1){0.4226}}\\\colorbox{white}{\makebox(10,1){0.4188}}\\\colorbox{white}{\makebox(10,1){0.4249}}\\\colorbox{white}{\makebox(10,1){0.4177}}\\\colorbox{white}{\makebox(10,1){0.4186}}\\\colorbox{white}{\makebox(10,1){0.4212}}\\\colorbox{white}{\makebox(10,1){0.4217}}\\\colorbox{white}{\makebox(10,1){0.4253}}\\\colorbox{white}{\makebox(10,1){0.4177}}\\\colorbox{white}{\makebox(10,1){0.4193}}\\\colorbox{white}{\makebox(10,1){0.4157}}\\\colorbox{white}{\makebox(10,1){0.4155}}\\\colorbox{white}{\makebox(10,1){0.4131}}\\\textbf{\textcolor{red}{0.4259}}\\\colorbox{white}{\makebox(10,1){0.4196}}\\\colorbox{white}{\makebox(10,1){0.4243}}}&
\makecell{\colorbox{white}{\makebox(10,1){0.7663}}\\\colorbox{green!30}{\makebox(10,1){0.7822}}\\\colorbox{green!30}{\makebox(10,1){0.7925}}\\\colorbox{white}{\makebox(10,1){0.7745}}\\\colorbox{green!30}{\makebox(10,1){0.7935}}\\\colorbox{green!30}{\makebox(10,1){0.7755}}\\\colorbox{green!30}{\makebox(10,1){0.7992}}\\\colorbox{green!30}{\makebox(10,1){0.7795}}\\\colorbox{green!30}{\makebox(10,1){0.7754}}\\\colorbox{green!30}{\makebox(10,1){0.7779}}\\\colorbox{green!30}{\makebox(10,1){0.7855}}\\\colorbox{green!30}{\makebox(10,1){0.7940}}\\\colorbox{white}{\makebox(10,1){0.7733}}\\\colorbox{white}{\makebox(10,1){0.7748}}\\\colorbox{white}{\makebox(10,1){0.7694}}\\\colorbox{white}{\makebox(10,1){0.7722}}\\\colorbox{white}{\makebox(10,1){0.7709}}\\\colorbox{green!30}{\makebox(10,1){\textbf{\textcolor{blue}{0.8004}}}}\\\colorbox{white}{\makebox(10,1){0.7750}}\\\colorbox{green!30}{\makebox(10,1){0.7967}}}&
\makecell{\colorbox{white}{\makebox(10,1){0.7027}}\\\colorbox{white}{\makebox(10,1){0.7174}}\\\colorbox{white}{\makebox(10,1){0.7268}}\\\colorbox{white}{\makebox(10,1){0.7101}}\\\colorbox{white}{\makebox(10,1){0.7283}}\\\colorbox{white}{\makebox(10,1){0.7120}}\\\colorbox{white}{\makebox(10,1){0.7327}}\\\colorbox{white}{\makebox(10,1){0.7150}}\\\colorbox{white}{\makebox(10,1){0.7118}}\\\colorbox{white}{\makebox(10,1){0.7145}}\\\colorbox{white}{\makebox(10,1){0.7205}}\\\colorbox{white}{\makebox(10,1){0.7280}}\\\colorbox{white}{\makebox(10,1){0.7098}}\\\colorbox{white}{\makebox(10,1){0.7096}}\\\colorbox{white}{\makebox(10,1){0.7049}}\\\colorbox{white}{\makebox(10,1){0.7074}}\\\colorbox{white}{\makebox(10,1){0.7067}}\\\textbf{\textcolor{red}{0.7338}}\\\colorbox{white}{\makebox(10,1){0.7119}}\\\colorbox{white}{\makebox(10,1){0.7308}}}&
\makecell{\colorbox{white}{\makebox(10,1){0.9140}}\\\colorbox{white}{\makebox(10,1){0.9137}}\\\colorbox{white}{\makebox(10,1){0.9122}}\\\colorbox{green!30}{\makebox(10,1){0.9142}}\\\colorbox{white}{\makebox(10,1){0.9127}}\\\colorbox{green!30}{\makebox(10,1){0.9140}}\\\colorbox{white}{\makebox(10,1){0.9140}}\\\colorbox{white}{\makebox(10,1){0.9133}}\\\colorbox{green!30}{\makebox(10,1){0.9144}}\\\colorbox{green!30}{\makebox(10,1){0.9144}}\\\colorbox{green!30}{\makebox(10,1){\textbf{\textcolor{blue}{0.9147}}}}\\\colorbox{green!30}{\makebox(10,1){0.9145}}\\\colorbox{green!30}{\makebox(10,1){0.9144}}\\\colorbox{white}{\makebox(10,1){0.9131}}\\\colorbox{white}{\makebox(10,1){0.9137}}\\\colorbox{white}{\makebox(10,1){0.9132}}\\\colorbox{white}{\makebox(10,1){0.9145}}\\\colorbox{white}{\makebox(10,1){0.9127}}\\\colorbox{green!30}{\makebox(10,1){\textbf{\textcolor{blue}{0.9147}}}}\\\colorbox{white}{\makebox(10,1){0.9134}}}&
\makecell{\colorbox{white}{\makebox(10,1){0.8618}}\\\colorbox{white}{\makebox(10,1){0.8614}}\\\colorbox{white}{\makebox(10,1){0.8589}}\\\colorbox{white}{\makebox(10,1){0.8622}}\\\colorbox{white}{\makebox(10,1){0.8597}}\\\colorbox{white}{\makebox(10,1){0.8619}}\\\colorbox{white}{\makebox(10,1){0.8612}}\\\colorbox{white}{\makebox(10,1){0.8608}}\\\colorbox{white}{\makebox(10,1){0.8625}}\\\colorbox{white}{\makebox(10,1){0.8624}}\\\textbf{\textcolor{red}{0.8628}}\\\colorbox{white}{\makebox(10,1){0.8621}}\\\colorbox{white}{\makebox(10,1){0.8625}}\\\colorbox{white}{\makebox(10,1){0.8603}}\\\colorbox{white}{\makebox(10,1){0.8615}}\\\colorbox{white}{\makebox(10,1){0.8610}}\\\colorbox{white}{\makebox(10,1){0.8622}}\\\colorbox{white}{\makebox(10,1){0.8596}}\\\textbf{\textcolor{red}{0.8628}}\\\colorbox{white}{\makebox(10,1){0.8602}}}&
\makecell{\colorbox{white}{\makebox(10,1){0.8497}}\\\colorbox{white}{\makebox(10,1){0.8495}}\\\colorbox{green!30}{\makebox(10,1){0.8508}}\\\colorbox{white}{\makebox(10,1){0.8495}}\\\colorbox{white}{\makebox(10,1){0.8487}}\\\colorbox{green!30}{\makebox(10,1){0.8501}}\\\colorbox{green!30}{\makebox(10,1){\textbf{\textcolor{blue}{0.8512}}}}\\\colorbox{white}{\makebox(10,1){0.8492}}\\\colorbox{white}{\makebox(10,1){0.8498}}\\\colorbox{green!30}{\makebox(10,1){0.8501}}\\\colorbox{green!30}{\makebox(10,1){0.8508}}\\\colorbox{green!30}{\makebox(10,1){0.8509}}\\\colorbox{white}{\makebox(10,1){0.8497}}\\\colorbox{green!30}{\makebox(10,1){0.8498}}\\\colorbox{white}{\makebox(10,1){0.8497}}\\\colorbox{white}{\makebox(10,1){0.8496}}\\\colorbox{white}{\makebox(10,1){0.8484}}\\\colorbox{white}{\makebox(10,1){0.8495}}\\\colorbox{white}{\makebox(10,1){0.8495}}\\\colorbox{green!30}{\makebox(10,1){0.8511}}}&
\makecell{\colorbox{white}{\makebox(10,1){0.8154}}\\\colorbox{white}{\makebox(10,1){0.8152}}\\\colorbox{white}{\makebox(10,1){0.8166}}\\\colorbox{white}{\makebox(10,1){0.8151}}\\\colorbox{white}{\makebox(10,1){0.8138}}\\\colorbox{white}{\makebox(10,1){0.8160}}\\\textbf{\textcolor{red}{0.8172}}\\\colorbox{white}{\makebox(10,1){0.8148}}\\\colorbox{white}{\makebox(10,1){0.8155}}\\\colorbox{white}{\makebox(10,1){0.8159}}\\\colorbox{white}{\makebox(10,1){0.8168}}\\\colorbox{white}{\makebox(10,1){0.8168}}\\\colorbox{white}{\makebox(10,1){0.8156}}\\\colorbox{white}{\makebox(10,1){0.8155}}\\\colorbox{white}{\makebox(10,1){0.8154}}\\\colorbox{white}{\makebox(10,1){0.8152}}\\\colorbox{white}{\makebox(10,1){0.8142}}\\\colorbox{white}{\makebox(10,1){0.8148}}\\\colorbox{white}{\makebox(10,1){0.8154}}\\\colorbox{white}{\makebox(10,1){0.8171}}}
\\ \bottomrule
\end{tabular}
}}
\label{tab:tests5}
\end{table}

The CC-CCII achieved better F-scores with the Grid Distortion, Rotate, and Shift Scale Rotate. Also, MedSeg had better F-scores in nine augmentations with the probability of 0.1 and seven augmentations with the probability of 0.2. Data augmentations did not improve the F-score in the Zenodo and Ricord1a datasets. The statistical analysis pointed out that the CC-CCII rejected the null hypotheses in seven data augmentations in both probabilities. Training for 100 epochs made the MedSeg achieve better F-scores in eleven data augmentations with probability 0.1 and seven data augmentations with probability 0.2. The results achieved in the MosMed dataset got worse when compared with the results presented in Table~\ref{tab:tests3}, with only seven data augmentations rejecting the null hypotheses. In the Ricord1a and Zenodo datasets, besides the average F-score of the data augmentations being very close to the average F-score without data augmentation, in the Ricord1a dataset, the null hypothesis was rejected in two data augmentations with probability 0.2, and in the Zenodo dataset, the null hypotheses were rejected in three data augmentations with probability 0.1 and two data augmentations with probability 0.2.

In the third evaluation, the architecture was trained for 100 epochs with a learning rate of 0.0001. The learning rate was divided by 10 every 25 epochs. Two probabilities of applying the data augmentation were evaluated: 0.1 and 0.2. As presented in Table~\ref{tab:tests5}, in the MosMed, the Shift Scale Rotate with the probability of 0.1 and the Rotate with the probability of 0.2 increased the F-score by 2\% compared with the baseline. Also, the MosMed achieved the best F-scores in nine augmentations with probability 0.1 and eight augmentations with the probability of 0.2, pointing this training configuration with the highest effectiveness for this dataset. Also, this training configuration showed significant effectiveness in the Zenodo dataset, which achieved the best F-scores in six augmentations with probability 0.1 and seven augmentations with probability 0.2. 

However, this training configuration did not significantly affect other datasets. The MedSeg achieved higher F-scores in only three augmentations with a probability of 0.1 and five augmentations with a probability of 0.2. Also, the CC-CCII and Ricord1a datasets did not achieve improvements with any data augmentation. Statistical analysis also was performed and, besides the average F-score of the data augmentations being very close to the baseline without data augmentation, the Zenodo and Ricord1a datasets achieved the highest number of data augmentations with the null hypotheses rejected. The MosMed was the dataset with the most promising results, with the null hypotheses rejected in twelve data augmentations. The CC-CCII and MedSeg had fewer data augmentations with the null hypotheses rejected compared with previews experiments, showing that this training configuration is not proper for data augmentations in these datasets.

%%%%%%%%%%%%%%%%%%%%%%%%%%%%%%%%%%%%%%%%%%%%%%%%%%%%%%%%%%%%%%%%%%%%%%%%%%%%%%%%
\section{Conclusion}

These three experiments demonstrated that, although necessary, the generic data augmentation techniques evaluated did not majorly improve the results in the COVID-19 segmentation problem. The MosMed achieved the most significant improvements with data augmentation, with the F-score being up to 2\% higher in comparison with baseline. This dataset was the most sensitive to data augmentation techniques due to its imbalance problem. Data augmentations also improved the CC-CCII and MedSeg dataset results, but it was necessary to train the network for more epochs, and the data augmentations only achieved 1\% of improvements in the F-score. The Ricord1a and Zenodo datasets were the most challenging and did not show improvements with data augmentations. Although they achieved statistical significance to reject the null hypotheses in many data augmentations, the average F-score slightly improved.

Another result of these data augmentation experiments is that spatial level transformations such as Elastic Transform, Flip, Grid Distortion, Piecewise Affine, Rotate, and Shift Scale Rotate were the operations that showed to be the most favorable data augmentations to this problem. These data augmentations improved the results in most of the experiments performed and thus are the most promising techniques for future experiments with data augmentation. The evaluation of more domain-specific data augmentation techniques was left for future works.

\section*{Acknowledgment}
The authors would like to thank the Coordination for the Improvement of Higher Education Personnel  (CAPES) for the PhD scholarship. We gratefully acknowledge the founders of the publicly available datasets, the support of NVIDIA Corporation with the donation of the GPUs used for this research and the C3SL-UFPR  group for the computational cluster infrastructure.

%\section{General Information}

%\section{Figures and Captions}\label{sec:figs}

%Figure and table captions should be centered if less than one line
%(Figure~\ref{fig:exampleFig1}), otherwise justified and indented by 0.8cm on
%both margins, as shown in Figure~\ref{fig:exampleFig2}. The caption font must
%be Helvetica, 10 point, boldface, with 6 points of space before and after each
%caption.

%\begin{figure}[ht]
%\centering
%\includegraphics[width=.5\textwidth]{fig1.jpg}
%\caption{A typical figure}
%\label{fig:exampleFig1}
%\end{figure}

%\begin{figure}[ht]
%\centering
%\includegraphics[width=.3\textwidth]{fig2.jpg}
%\caption{This figure is an example of a figure caption taking more than one
%  line and justified considering margins mentioned in Section~\ref{sec:figs}.}
%\label{fig:exampleFig2}
%\end{figure}

%In tables, try to avoid the use of colored or shaded backgrounds, and avoid
%thick, doubled, or unnecessary framing lines. When reporting empirical data,
%do not use more decimal digits than warranted by their precision and
%reproducibility. Table caption must be placed before the table (see Table 1)
%and the font used must also be Helvetica, 10 point, boldface, with 6 points of
%space before and after each caption.

%\begin{table}[ht]
%\centering
%\caption{Variables to be considered on the evaluation of interaction
%  techniques}
%\label{tab:exTable1}
%\includegraphics[width=.7\textwidth]{table.jpg}
%\end{table}

\bibliographystyle{sbc}
\bibliography{main}

\end{document}